\documentclass[sigconf, nonacm]{acmart}

\usepackage{amsmath}
\usepackage{algorithm}
\usepackage{algpseudocode}
\usepackage{graphicx}
\usepackage{subcaption}
\usepackage{colortbl}
\usepackage{balance}

\begin{document}

\title{StreamEnsemble: Predictive Queries over Spatiotemporal Streaming Data}

\author{Anderson Silva}
\affiliation{
 \institution{LNCC}
 \city{Petrópolis}
 \country{Brazil}
}
\email{achaves@lncc.br}

\author{Eduardo Ogasawara}
\affiliation{
 \institution{CEFET/RJ}
 \city{Rio de Janeiro}
 \country{Brazil}
}
\email{eogasawara@ieee.org}

\author{Patrick Valduriez}
\affiliation{
 \institution{Inria, University of Montpellier, CNRS, LIRMM}
 \city{Montpellier}
 \country{France}}
 \affiliation{
 \institution{LNCC}
 \city{Petrópolis}
 \country{Brazil}
}
\email{patrick.valduriez@inria.fr}

\author{Fabio Porto}
\affiliation{
 \institution{LNCC}
 \city{Petrópolis}
 \country{Brazil}
}
\email{fporto@lncc.br}

\renewcommand{\shortauthors}{Silva et al.}

\begin{abstract}

Predictive queries over spatiotemporal (ST) stream data pose significant data processing and analysis challenges. ST data streams involve a set of time series whose data distributions may vary in space and time, exhibiting multiple distinct patterns.
In this context, assuming a single machine learning model would adequately handle such variations is likely to lead to failure. To address this challenge, we propose StreamEnsemble, a novel approach to predictive queries over ST data that dynamically selects and allocates Machine Learning models according to the underlying time series distributions and model characteristics. Our experimental evaluation reveals that this method markedly outperforms traditional ensemble methods and single model approaches in terms of accuracy and time, demonstrating a significant reduction in prediction error of more than 10 times compared to traditional approaches. 

\end{abstract}




\maketitle

\section{Introduction}

Spatiotemporal (ST) data is critical for predictive analysis in many applications, such as temperature/precipitation forecasting \citep{ravuri_skilful_2021, chattopadhyay_predicting_2020, das_probabilistic_2015, bonnet_precipitation_2020}, epidemic propagation \citep{yu_spatio-temporal_2023, matsubara_funnel_2014}, neuroscience \citep{kleesiek_deep_2016, nie_3d_2016, wen_deep_2018}, traffic flow prediction \citep{huang_deep_2014, sun_dxnat_2017}, and social sciences \citep{afyouni_deep-eware_2022, sagl_social_2012}. ST data capture both location and time information of measured events and are frequently delivered in high-speed, high-volume data streams, demanding efficient processing techniques capable of handling their continuous influx. In these scenarios, \textit{predictive queries} apply predictive models over the incoming ST data streams to make forecasts based on identified patterns and trends. The primary role of these models is to be able to understand connections and interdependencies within the data to enable accurate forecasts based on patterns, which, in the case of ST data streams, are particularly complex and subject to dynamic fluctuations \citep{hamdi_spatiotemporal_2022}. 

Typically, a range of available trained machine learning (ML) models are considered when composing a predictive query. Each model is trained on a different data region and learns data patterns from a variety of time series. In these scenarios, the correct choice of a single appropriate model to be used by a query is challenging since determining which model best suits the series for a given region can be difficult. Furthermore, each model specializes in different data distributions, with strengths and weaknesses. Thus, combining a set of models requires a compelling strategy, allowing different aspects of the data to be captured and providing diverse perspectives and complementary answers to enhance the overall accuracy. However, selecting and combining these models effectively is hard since the data and models' characteristics need to be considered. This challenge can be expressed as the following model selection problem: Given a predictive query over an ST data stream and a set of ST predictors, how can we select and combine the predictors to minimize the total inference cost? Here, ``cost'' can refer to accuracy, inference time, or a combination of both.

Traditional model ensembles, where multiple models are used complementarily, have been demonstrated as an effective approach in various applications \citep{sagi_ensemble_2018, dong_survey_2020}, including over data streams \citep{gomes_survey_2017, krawczyk_ensemble_2017}. However, this approach may incur the following problems: (1) high execution time if the available base models are complex since it requires running all of them to post-process and combine their results; (2) poor prediction accuracy at ST data, since time series data distribution may significantly vary according to their distribution along the spatial and temporal region of the input data, thus becoming hard to determine the weights for each model and compose the ensemble accordingly. 

Alternatively, a globally trained model representing multiple diverse time series occurring in its training data and thus capturing a wider range of underlying data distributions could be a compelling strategy. However, such a model is typically not available since (1) it is difficult to ensure an adequate amount of training data to encompass all patterns present in every region and (2) coping with the increased complexity and computational demands for training a model that learns these multiple patterns across different regions and temporal periods may not be feasible with the available resources.

\begin{figure}[!ht]
	\centering	
 \includegraphics[width=1.0\columnwidth]{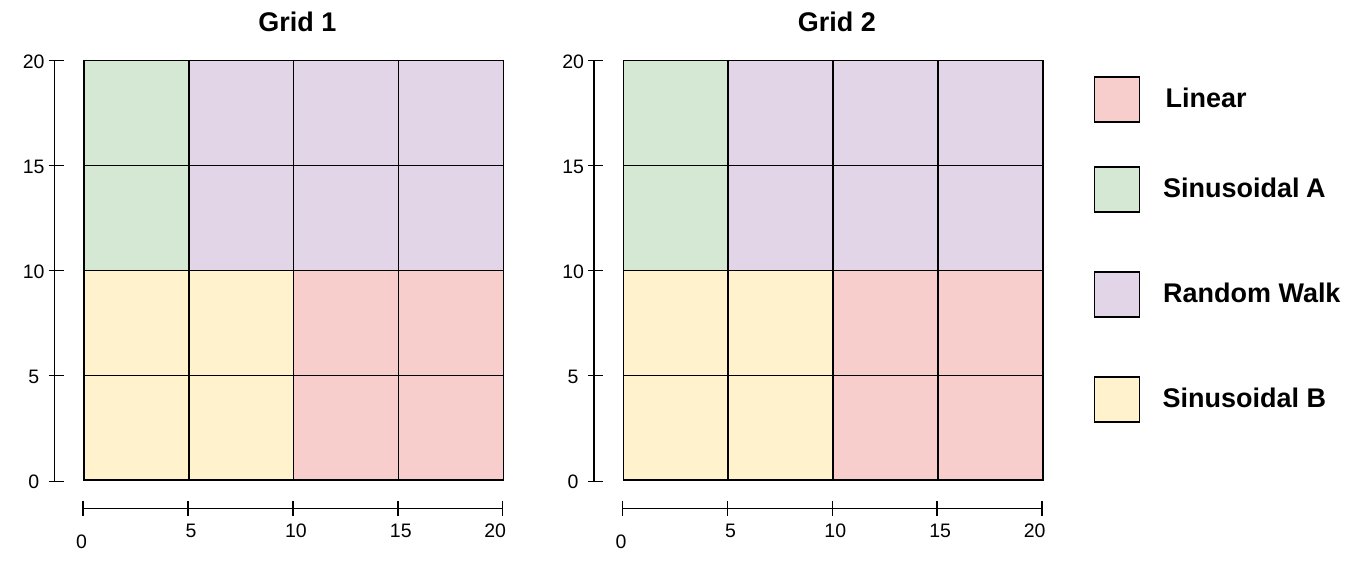} 
 \caption{Two 20x20 time series grids composed of 16 regions. All time series within a region share a common pattern}
	\label{fig_motivation}
\end{figure}

To illustrate our problem, consider the following (synthetic) experiment using a carefully constructed dataset. The dataset comprises 16 regions of time series, each region of length 5x5, all forming a structured time series grid as shown in Figure \ref{fig_motivation}. Within each region, all time series share a common pattern, which could be linear, sinusoidal with two different amplitudes and frequencies, or random walk. Two different grids are defined with different pattern configurations. 

In this scenario, we trained a unique model for each distinct pattern in the dataset and a global model of similar complexity on a training set that includes all patterns collectively. Then, we implemented distinct model allocation strategies to assess their impact on prediction accuracy: Random Allocation, where models are allocated randomly across different spatial areas; Best Fit Allocation, where models are allocated based on their compatibility with the prevalent pattern in each spatial region; and Global Model Allocation, assigning to all spatial areas the global model. We evaluated the performance of each allocation strategy for both Grid 1 and 2 by assessing the average prediction error for the corresponding models. To simulate the impact of a change in the data patterns, we evaluated Grid 2 with a static best-fit allocation strategy, where the model allocation chosen for Grid 1 is also used for Grid 2, and a dynamic best-fit allocation, where models are allocated according to the patterns for each grid.

\begin{figure}[!ht]
	\centering	
 \includegraphics[width=1\columnwidth]{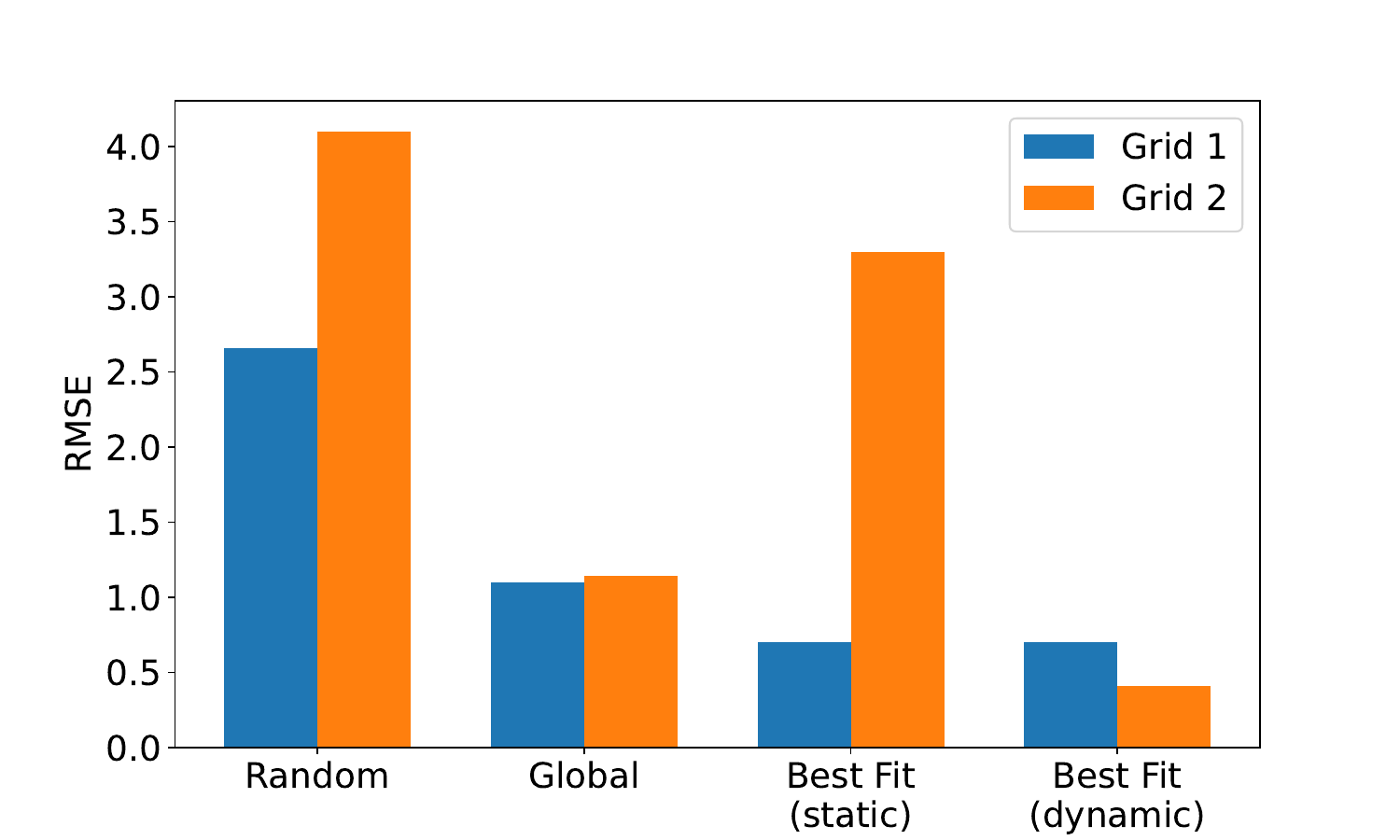} 
	\caption{Impact of Allocation Strategies on Prediction Accuracy, using different model allocation strategies.}
	\label{fig_baseline_results}
\end{figure}

The results are depicted in Figure \ref{fig_baseline_results}. Within Grid 1, the Best Fit Allocation strategy performs better than the Random and Global Model Allocation strategies. This shows that a carefully chosen set of models is necessary to compute predictions on input stream coming from  spatial regions with varying data patterns. Within Grid 2, a static Best Fit Allocation strategy exhibits suboptimal performance since it considers the same model allocation chosen for Grid 1, while the dynamic Best Fit allocation presents accurate results since models are chosen specifically according to the patterns for Grid 2. This experiment underscores that a solution composed of carefully selected specialist models aligned with the prevailing pattern in the target data is more efficient than alternative approaches. 

To address the model selection and combination problem, we propose a method where decisions are made based on the data distribution of the incoming time series windows and the model generalization capability. Our approach involves spatially clustering similar series based on their patterns and characteristics. Additionally, we employ a generalization error estimation technique to identify the most suitable models for each specific target, which are then systematically chosen to form the ensemble. Our solution proposes adapting the model ensemble to changing data patterns, seeking to optimize the forecasting accuracy to data distribution variations in ST streaming scenarios.

By systematically choosing the most suitable models for each specific target, StreamEnsemble achieves performance comparable to the optimal solution provided by isolated models. Furthermore, it 
outperforms the globally-trained model by a substantial margin, in some cases exceeding a 10x reduction in prediction error and highlighting the limitations of a Single-model approach for heterogeneous ST data streams.

In summary, our contributions are:

\begin{itemize}
 \item A method for online selection and allocation of models over ST data based on the training data distribution and each model's generalization error that considers the variability of the target data distributions in space and time.

 \item An empirical validation of this method through a comprehensive set of experiments that assess its efficiency on meteorology datasets and a set of 24 ST predictors (STPs). 

 \item An exploration of alternative approaches for implementing the proposed methodology, including the representation of time series, clustering techniques, and other fundamental components. 
\end{itemize}

This paper is organized as follows. The next section introduces a practical example that illustrates the presented problem. Section \ref{sec_background} introduces essential background concepts related to ST data stream processing and model ensembling, and Section \ref{sec_problem_definition} formally defines the problem.
Section \ref{sec_stream_ensemble} describes our method. Our experimental evaluation is presented in Section \ref{sec_experimental_evaluation}, where we also discuss the practical implications of our approach. In Section \ref{sec_related_work}, we present related works to contextualize our contributions. Finally, in Section \ref{sec_conclusion}, we summarize the main findings of this paper and outline potential directions for future research and development.

\section{Motivating Example}
\label{sec_example}

Consider a climate monitoring system that preemptively notifies public agencies about regions that are potentially subject to meteorological events of interest. Various heterogeneous sensor devices, such as temperature monitoring stations or pluviometrical radars, produce sequences of measurements at specific coordinates. By processing the data from the sensors, these sequences are used to construct data frames at fixed time intervals, forming the basis for subsequent analysis.

In this context, a predictive query specifies a subregion of interest where operations must be performed to anticipate future meteorological events. A collection of ML models is available to compose the query and produce accurate forecasts. These models, independently trained on different geographical regions, learned unique patterns, each passing through dedicated life cycles encompassing data acquisition, training, and hyperparameter optimization. 

\begin{figure}[!ht]
	\centering	
 \includegraphics[width=1\columnwidth]{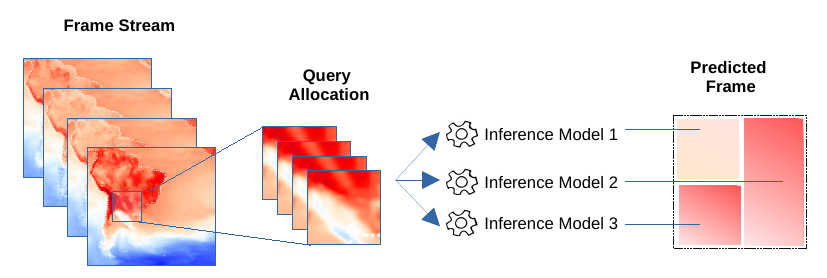} 
	\caption{ST Model Ensemble defined over query region}
	\label{fig_problem-figure}
\end{figure}

Figure \ref{fig_problem-figure} illustrates the example. In this context, the predictive query is an ongoing data prediction operator applied to a series of incoming data windows to provide the most recent insights. Given the data's dynamic nature, the challenge is to adaptively compose an ensemble of ML models based on the evolving characteristics of the ST time series. A potential solution needs to continuously monitor the data streams, adapt to changes in data distribution and periodically recompose the ensemble by dynamically selecting and combining the most appropriate models to accurately predict the changing environmental conditions across different geographical regions. Additional factors should also be considered for this model allocation, such as the models' generalization capabilities and the variable dimensionality of their input data, which corresponds to spatial regions of varying sizes.

\section{Background}
\label{sec_background}
This section provides a comprehensive overview of the fundamental ST stream data analysis related to our solution.

\subsection{Stream Data}
\label{sec_stream_data}

\textit{Stream data} is a continuous and potentially unbounded flow of generated data that must typically be processed in real-time or near real-time. In general, the main challenges when dealing with stream data are (1) limited computing resources in terms of memory, bandwidth, and time since data is generated continuously, potentially unbounded; (2) predictions must be made with a short delay, which can make impossible the labeling of data items on time and (3) the streams data distribution may change unpredictably over time, incurring in poor predictions and incorrect decisions from the models \citep{lu_learning_2019}. 

In the context of data analysis, \textit{ST data streams}, produced by sources such as GPS devices, satellite imagery, and weather monitoring stations, introduce an extra layer of complexity: the spatial dimension. In these data streams, each data point is typically associated with a set of spatial coordinates (e.g., latitude and longitude) and a timestamp. Another characteristic is the existence of dependencies among time series due to the spatial dimensions. 

\subsection{Time Series Clustering}

To overcome the limitations of clustering techniques over stream data, \textit{Stream Clustering} algorithms have been proposed. The main objective of these methods is to continually update existing clusters, assimilate emerging patterns, and incrementally discard outdated ones. This involves optimizing the number and location of clusters to represent the underlying data best and extract relevant information without storing and re-evaluating all observations. Preserving time-based patterns by representing the incoming data as time series is useful in accounting for temporal dependencies. Nevertheless, \textit{time series clustering} faces challenges that impact the efficiency and effectiveness of the clustering process:

\begin{itemize}
 \item Data Size and Storage: To perform efficiently, the clustering task often requires repeatedly accessing the whole time series data for multiple iterations. However, storing all historical data for each time series in memory is often impractical since it may exceed memory capacity. 
 \item High Dimensionality: If we assume each point in the series is a feature, time series often exhibit high dimensionality, degrading the accuracy of many clustering algorithms and slowing the clustering process.
 
 \item Choice of a similarity measure: Time series similarity matching calculates the similarity among the entire time series using a similarity measure. Similarity between time series is a well-studied topic in literature, and many different metrics have been proposed \citep{bellman_adaptive_1958, moller-levet_fuzzy_2003, warren_liao_clustering_2005}. However, comparing different time series can be challenging because data can be complex due to inherent noise, outliers, and shifts \citep{aghabozorgi_time-series_2015}. These factors must be considered when selecting the most suitable similarity measure for a given dataset and analytical task.
\end{itemize}

\subsection{ST Series Forecasting}

Sequence Forecasting is the task of predicting the most likely sequence of observations in the future, given the previous observations. In many forecasting problems, ST information is very useful for increasing prediction accuracy. In temperature prediction, for example, the latitudes are also essential for predicting future values besides the historical sequence of temperature measurements. ST sequences offer a useful representation of the underlying dynamic system by preserving spatial and temporal correlations among observations (like videos). These complex relationships, however, represent big challenges, especially for forecasting extensive sequences \citep{shi_machine_2018}.

ST forecasting can be categorized based on the grid structure, distinguishing between regular and irregular grids. In cases where it is possible to collect data from extensive regions using densely distributed sensors or monitor large geographic areas using radar or satellite systems \citep{ravuri_skilful_2021}, the regular grid representation is typically straightforward. Conversely, when addressing issues that involve limited monitoring points (e.g., disease outbreak prediction problems), it is possible to derive a regular grid from an irregular one by estimating values for the remaining positions within the irregular grid. 

In typical sequence forecasting methods, it is often assumed that each sequence is not influenced by or related to others, and all have the same statistical properties - in other words, they are independent and identically distributed (i.i.d.) \citep{atluri_spatio-temporal_2018, eklund_cluster_2016}. However, in ST data streams, the nature of the data distribution tends to vary according to time and space. For example, climate data such as temperature or rainfall presents cyclic patterns due to changing seasons. Geographical features such as mountainous or coastal regions may also introduce statistical differences. Observations at different points can also exhibit spatial dependence, meaning that a sequence behavior at one location may correlate with the values at nearby locations. Failing to consider these factors could dramatically deteriorate the selected method's performance.

Formally, on a regular grid, ST predictors try to approximate the function:

$$
\arg \!\!\!\!\!\!\!\! \max_{X_{t+1}, ..., X_{t+k}} \!\!\!\! p(X_{t+1}, X_{t+2}, ..., X_{t+k} | X_{t-j+1}, X_{t-j+2}, ..., X_t)
$$

where $p(x)$ is a conditional probability and each $X_i \in R^{P \times M \times N}$ represents the data values on a grid of $M$ rows, $N$ columns and $P$ measurements at time $i$. Based on $k = 1$ or $k > 1$, the forecasting can also be classified as single-step or multi-step. Compared with single-step forecasting, learning a model for multi-step forecasting requires more elaborate models since the forecasting output $X_{t+1}, ..., X_{t+k}$ is a sequence with non-i.i.d elements. 

\section{Problem Definition}
\label{sec_problem_definition}

We consider a dynamic system over a spatial region represented by an $M \times N$ grid consisting of $M$ rows and $N$ columns. Inside each cell in the grid, a measurement varies over time. Thus, the observation at any time $t$ can be represented by an observation matrix, or frame, $X_t \in R^{M \times N}$ where $R$ denotes the domain of the observed features. For the sake of simplicity, we present our scenario by focusing on forecasting only the next value of an univariate series using the past $n$ measurements through data windows of fixed size $n$. Thus, in a given period, we can obtain an ST sequence of $n$ observations $X_{t-n+1}, X_{t-n+2}, \dots, X_{t}$, which we call a data window at $W_t$.

To compute $X_{t+1}$, a set of ST Predictors (STPs) $M = \{m_1, m_2, \dots, m_s\}$ is available. It is worth noting that the dimensions of each model's input frames may vary across different models. This flexibility allows us to adapt the models to various spatial configurations and capture relevant information effectively. Given a ST query $Q$ over $W_t$, our problem involves determining a partitioning $P$ of the query space and a subset of models $M' \subseteq M$, so that: $P = \bigcup_{i=1}^n P_i, P_i \cap P_j = \emptyset, \quad 1 \leq i, j \leq n, \quad i \neq j$, then finding the optimal allocation $A_{t} = (P_i, m_j)_{t}$ of STPs at each timestamp $t$. The execution of the models $M'$ from allocation $A_{t}$ produces a ST prediction $\hat{X}_{t+i}$ that satisfies the region defined by $Q$ and has an execution cost that minimizes $(\hat{X}_{t+i} - X_{t+i})$, so that:

(i) $\forall P_i \in P, \exists$ model $m_j \in M^{\prime}$ such that $A\left(R_i, m_j\right)$ \\

(ii) $A\left(P_j, m_i\right) \wedge A\left(P_j, m_k\right)$ if only if $i=k$

Under these definitions, while only one model per partition is allocated, a single query can span multiple partitions.

\section{Stream Ensemble}
\label{sec_stream_ensemble}

\begin{figure}[!ht]
	\centering	
 \includegraphics[width=1\columnwidth]{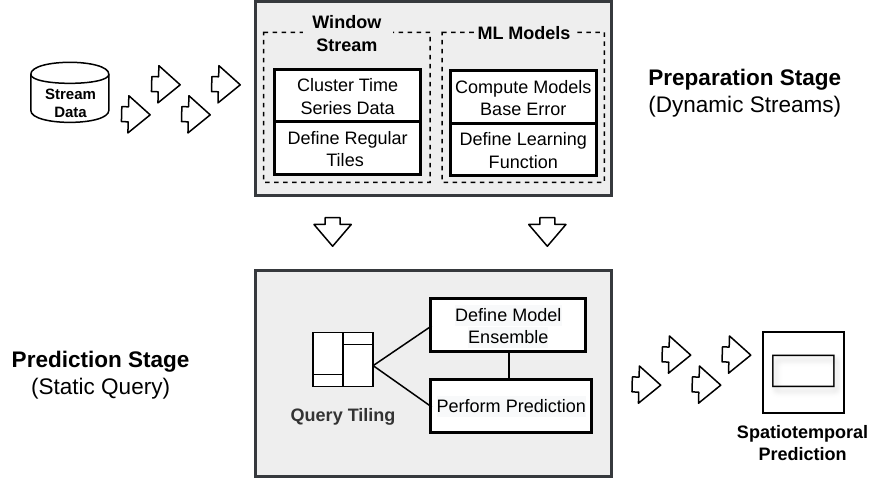} 
	\caption{StreamEnsemble}
	\label{fig_stream-ensemble}
\end{figure}

To address the aforementioned challenges in dealing with the selection of ML models over ST predictive queries, we propose a novel method that is composed of three distinct but interconnected steps: (1) creating an error estimation function for each available model, (2) partitioning the query space into subregions of similar data distributions, and (3) composing a model ensemble based on smallest estimated execution cost. This section provides an in-depth explanation of each of these steps. Figure \ref{fig_stream-ensemble} illustrates the main steps of the method. Steps (1) and (2) are addressed in the preparation stage, while step (3) is addressed in the prediction stage. 

\subsection{Estimating Models Generalization Error}
\label{sec_estimating_models}

The initial step is a method to estimate the execution error for each candidate model. The goal is to develop a function that quantifies how well a model performs on unseen data, an essential step for ensemble selection. The estimation process is unique in that it considers the space in which the model has been trained, i.e., its training data and generalization capabilities.

For each candidate model, we compute the model's error on its training dataset $T$. Gaussian noise with increasing variance is added to these training data points to inject the element of uncertainty. This operation creates a sequence of altered datasets, namely ${T_0, \ldots, T_n}$, where each dataset $T_i$ has been subjected to $i$ iterations of Gaussian noise, incrementally increasing its variance. 

Subsequently, the strategy measures the dissimilarity between $T$ and each $T_i$. This dissimilarity is quantified using the Dynamic Time Warping (DTW) function, which is applied over the centroid series of each dataset. DTW measures the similarity between the two time series, considering their temporal alignment and value differences. A regression model of the form $error = F(dist_{i})$ is fitted using cross-validation from each pair DTW distance and model error. This regression model is referred to as the Error Estimation Function. It captures the relationship between the distance in the feature space and the model's error, enabling the estimation of the model's accuracy in unseen data regions.

\subsection{Query Partitioning}
\label{sec_query_partitioning}

The next step involves partitioning the query space. ST data often contains many underlying patterns and behaviors that vary across geographical regions and periods. For instance, when analyzing traffic data, one subset could consist of congested urban areas, while another might concentrate on open highways. Similarly, certain geographical areas might exhibit unique climate patterns or pollution levels when monitoring the environment. By partitioning the data stream, we can isolate these patterns within subregions. Defining these subregions enables us to choose the most appropriate model for each area, i.e., one that could capture the particular phenomena observed in that subset. 

Partitioning and selecting a single best-suited model for each region also contributes to efficient resource allocation. Instead of applying all models to the entire dataset, which can be computationally demanding, resources can be allocated individually to a single model per partition.

The second step of our method focuses on partitioning the incoming ST data stream into subregions of similar behaviors. This step involves three key sub-processes: representation, clustering, and tiling.

\subsubsection{Time series representation}

Time series often possess high dimensionality; each data point is a time-stamped observation. These high-dimensional representations can be computationally intensive and challenging to analyze effectively. Our proposed method first represents the time series on each data window as a lower-dimensional vector. This step aims to reduce the time series length without significant information loss. A lower-dimensional representation enables the subsequent clustering algorithm to process data more efficiently, mitigating the risk of overfitting and ultimately leading to more robust and accurate clusters.
Furthermore, time series frequently presents inherent noise that can arise from various sources, such as measurement errors, outliers, or inherent variability. These elements can lead to spurious clusters or misinterpretation of temporal patterns. By reducing the dimensionality of the data, we intend to extract and emphasize the series' most relevant and informative components. 

We explore two different functions for dimensionality reduction: the Generalized Lambda Distribution (GLD) \citep{ramberg_approximate_1974} and Parcorr \cite{yagoubi_parcorr_2018}, a parallel incremental random vector/sketching approach.

The \textit{Generalized Lambda Distribution} is a statistical distribution function that offers a flexible and versatile framework for modeling a wide range of probability distributions. For this reason, it is particularly useful in representing time series. The GLD is defined by four parameters ($\lambda_1, \lambda_2, \lambda_3, \lambda_4$) that govern its shape and properties: the location parameter ($\lambda_1$) shifts the distribution along the x-axis (i.e., the mean), the scale parameter ($\lambda_2$) controls its spread, the shape parameter ($\lambda_3$) influences skewness and kurtosis, and the tail parameter ($\lambda_4$) determines tail thickness, impacting the probability of extreme values in the distribution. The GLD can represent time series by fitting the distribution to the data's empirical probability density function (PDF). The goal is to find the GLD parameters that best approximate the observed data distribution. Statistical estimation techniques can be used to determine the GLD parameters that minimize the discrepancy between the observed data and the fitted GLD \citep{fournier_estimating_2007, staden_method_2009, bergevin_analysis_1993}.

\textit{ParCorr} is an efficient method designed for identifying similar time series pairs within sliding windows of data streams, using either Euclidean distance or Pearson correlation as a similarity metric. The fundamental idea is to compute the dot product between each normalized time series within a specified window size $w$ and a set of $b$ random vectors. This process results in a \emph{sketch} of the time series for that particular window. For two time series $t_i$ and $t_j$, it is mathematically demonstrated that the distance $||sketch(t_i) - sketch(t_j)||$ is a good approximation of $||t_i - t_j||$ provided the dimensionality of the sketches $b$ is large enough. Comparing these sketches allows for identifying time series close in the sketch space, which is computationally much more efficient, especially when $w$ is much larger than $b$.

\subsubsection{Time series clustering}

Once each series representation is well defined, this step aims to group similar time series, enabling the identification of homogeneous allocation regions for associating each ML model. This process optimizes the allocation of models to different data partitions. We propose three distinct interchangeable clustering strategies, depending on the use cases and computational requirements:

\textbf{Static clustering} utilizes a predefined clustering algorithm, such as k-means, to group time series based on an initial, fixed-length time window. Once the clustering is defined, it is never updated. This method is particularly suitable when the cluster allocations for each region remain relatively constant over time, and the overhead of re-executing the clustering step is impractical. We choose the silhouette test to determine the optimal number of clusters for a given dataset and time window, quantifying how similar each series is to its corresponding cluster compared to others. 

\textbf{Dynamic Clustering}, in contrast to static clustering, re-executes the clustering step for each time window. Although this approach may be more computationally expensive due to its iterative nature, it is well-suited for scenarios where the time series within regions exhibit dynamic characteristics. It allows the selected models to be adapted to evolving data distributions by continuously reassessing cluster assignments. If the execution time constraint of the problem allows it, the number of clusters can also be determined dynamically based on clustering metrics such as the silhouette score. By reassessing the optimal number of clusters for each time window, our proposed strategy ensures that the cluster assignments remain in sync with the changing data distribution.

\textbf{Stream clustering}: In cases where real-time adaptability is necessary, the approach of stream clustering can be more adequate. Stream clustering is particularly valuable when handling unstable clusterings and high-velocity streaming data and situations where computational efficiency is required, providing continuous adaptation to gradually changing clusterings. This approach utilizes stream clustering algorithms that can handle large datasets incrementally with limited memory usage. Specifically, we implement the BIRCH algorithm \citep{zhang_birch_1996}, a highly efficient clustering algorithm \citep{lorbeer_variations_2018}. Its fundamental idea is to construct a hierarchical structure of clusters, allowing for efficient retrieval and updating of cluster assignments as new data arrives. 

By incorporating these three clustering strategies, the algorithm offers the flexibility to adapt to diverse data characteristics and computational constraints, whether data clusters remain relatively stable over time, evolve dynamically, or demand real-time adaptability.

\subsubsection{Tiling}
\label{sec_tiling}

The tiling step ensures that the input to multidimensional ML models, such as convolutional models, is a regular and well-structured frame that preserves the underlying data distributions revealed in the clustering step. This step addresses the geometrical challenges that arise from irregular clustering outputs and noisy datasets. It transforms the irregular clustering results into regular, connected regions known as tiles, which serve as input to the ML models.

This step is necessary because of the geometric misalignment between the input frames, which are typically regular matrices, and the often irregular clustering results. Irregular clusters can result from the data's inherent noise or the non-uniform distribution of data patterns. To consider these factors, the tiling process aims to produce connected, regular regions for the ML models to analyze.

Two main strategies for tiling are employed, each with specific characteristics:

\textbf{Bottom-Up Tiling:}
In the bottom-up tiling strategy, an adaptation inspired by the YOLO (You Only Look Once) algorithm is employed. This strategy initially assigns a tile to a single time series or data point not associated with any other tile. It then systematically grows the tile's area in a given direction (up, down, left, or right), expanding its boundaries. When considering the addition of series from a particular dimension, the algorithm calculates the potential impact on the tile's purity, a measure of how homogeneous the data within the tile is regarding cluster assignments.

The algorithm begins with a defined purity threshold, determined at the start of the tiling process. If expanding the tile in a specific direction would result in a purity level below the threshold, the algorithm explores other spatial dimensions, effectively avoiding adding data that would reduce the tile's purity. Conversely, if adding data maintains or enhances the tile's purity above the threshold, the algorithm expands the tile in that direction.

This process continues iteratively, expanding the tile's area while ensuring that the tile maintains a high degree of homogeneity in cluster assignments. When the tile cannot be expanded, the process is repeated for a new tile. 

\textbf{Top-Down Tiling:}
In contrast to bottom-up tiling, the top-down tiling strategy utilizes a partitioning approach to transform the query space into square subregions iteratively. The quad-tree data structure is employed to partition the space into square regions, dividing hierarchically each region into four equal-sized squares or quadrants, allowing for a recursive partitioning process. The algorithm keeps subdividing each quadrant until the purity rate satisfies (exceeds) the specified minimum threshold.

In both tiling strategies, the primary goal is to create regular, connected regions with a minimum percentage of series belonging to a single cluster, ultimately optimizing the input data structure for the multidimensional ML models. These structured regions, the tiles, geometrically assist the frame allocation to the selected ML models, enabling the execution of ST predictors that generally require the input to consist of a regular structure.

The choice between top-down and bottom-up tiling strategies depends on the specific characteristics of the data and the application's requirements. Top-down tiling tends to produce larger, more coherent regions, while bottom-up tiling offers finer-grained partitions but may produce larger tiles.

\subsection{Model Ensemble Composition}

After partitioning the query area into tiles, the method finally proceeds with the selection and allocation of the most suitable ML model for each tile. The guiding principle in this step is to minimize the estimated execution cost by utilizing the Error Estimation Function (EEF), defined in Section \ref{sec_estimating_models}. The EEF estimates the model's error for each tile based on a representative time series that characterizes the data distribution and behavior within that tile.

Using the EEFs defined for each candidate ML model, an error estimation for each model can be calculated for each tile. The reference time series for each tile is identified as the medoid, which serves as the central, most representative time series within the tile. This process involves calculating the estimated error for each pair of tiles and model. As a key feature, this step is not computationally expensive, as it necessitates no more than the EEF computations for each pair (tile, model). Subsequently, the model with the lowest estimated error for each tile is selected as the best predictor for that specific region.

Once the best-suited model has been selected for a particular tile, the method addresses the spatial dimension misalignment that often exists between the tile and the model's input window. The tile's spatial dimensions may be either larger or smaller than those of the model's input window. The method adapts accordingly.

If the tile is larger than the model's input window, the model is allocated multiple times within the tile, with disjoint allocations. This approach effectively fits the model into the larger tile, allowing it to analyze different tile segments without overlap, as depicted in Figure \ref{fig_problem-figure}. Each model allocation processes a portion of the tile data.

If the model's input window is larger than the tile, or if the tile dimensions are larger but not multiples of the model's input window, the composition algorithm duplicates the last columns or rows of the tile data to match the model's input window dimensions. This ensures that the model receives a regular input format for processing.

For each tile, the selected ML model is executed independently. The final step involves recombining the predictions generated by each model for each tile. This process is straightforward and involves the following steps:

\begin{enumerate}
 \item For each model's output, select the portion of the output corresponding to the data within each tile. 
 \item Combine these model predictions into the output window, aligning them correctly to reconstruct the query space according to the spatial input dimensions.
\end{enumerate}

By selecting the most suitable models for each tile based on the error estimation function and efficiently aligning the tile and model dimensions, the resulting predicted frame represents the prediction of the most suited ensemble over continuous queries on streaming data, contributing to improved accuracy in query processing.

While the outlined process is suitable for predicting a single frame based on each data window, it should be noted that for an interval $k$ query, the process may be repeated $k$ times. The procedure $StreamEnsemble(Q, M)$ where $Q$ represents the query and $M$ denotes a collection of models. illustrates each step of the method. 

\begin{algorithm}[!ht]
\small
\begin{algorithmic}[1] 
\Procedure{StreamEnsemble}{$Q, M$}

\For{$m \in M $}
 \State{$Cost_m \gets$ define cost function($m$)}
\EndFor 
\State{$t_c \gets 0$}
\While{data stream is active}
 \State{read data window $W_{t_c} = (X_{t_c-j+1}, ..., X_{t_c})$;}
 \State{Tiles $T$ = partition query space($Q, W_{t}$)}
 \State{Ensemble $E$ = compose\_ensemble($T, Cost$)}
 \State{perform\_continuous\_query($Q, E$) }
 \State {$t_c \gets t_c + 1$}
\EndWhile 
\EndProcedure
\end{algorithmic} 
\label{alg:StreamEnsemble}
\end{algorithm} 

\section{Experimental Evaluation}
\label{sec_experimental_evaluation}

In this section, we analyze algorithmic performance within our strategy. We explore the choices available for each component of our approach and compare their effectiveness and suitability. 
To better evaluate the approach's contribution, we utilize simple methods that do not rely on sophisticated feature engineering, intricate deep learning architectures, or hyperparameter optimization.
Additionally, we compare our proposed solution and the baseline approach, which does not incorporate an ensemble, to evaluate the impact of our strategy.

\subsection{Experimental Setup}

In the following, we present the hardware and software configurations and the data workload and models utilized. 

\subsubsection{Environment} All the experiments were conducted on a Dell PowerEdge R730 server equipped with dual Intel Xeon E5-2690 v3 processors clocked at 2.60GHz and 768GB of RAM. The server operates on a Linux CentOS 7.7.1908 platform with a kernel version 3.10.0-1062.4.3.e17.x86 64. The models were trained and tested using an NVIDIA Pascal P100 GPU with 16GB of dedicated RAM. The algorithms and all related code were implemented in Python, leveraging popular libraries such as NumPy, SciPy, Scikit-learn, and custom-developed functions written to our specific research needs. 

\subsubsection{Dataset}
The CFSR (Climate Forecast System Reanalysis) dataset includes temperature data collected over 25 years measured every six hours, providing a robust variety of climate and weather patterns. This dataset is particularly suitable for our research as it offers a rich source of ST data, allowing us to explore our method's performance in real-world temperature data analysis. We used a subset of this dataset, a region of 141 x 153 cells corresponding to South America, in a resolution of 0.5° x 0.5° degrees latitude and longitude. 

\subsubsection{ML Models Configuration} In our experiments, we employed a diverse range of models, each characterized by specific parameters while considering three key dimensions of model variability: input window size, training timestamps, and training regions. The following details describe the configurations of these models on each dimension.

\textbf{Input Size}: To explore the impact of input window size on the ensemble composition, we considered four different input window lengths: 1x1, 3x3, 5x5, and 7x7. Models configured with larger window sizes gain a broader spatial scope, which enhances their ability to identify overarching patterns within the input data. Also, models with smaller input regions must be invoked fewer times when executed over a large tile when chosen to compose the ensemble. There is a trade-off when the tile size is smaller than the models' input, necessitating artificial data synthesis to complete the models' input window. In scenarios dominated by smaller tiles, models with compact window sizes are the more fitting choice since they do not require the synthesis of additional data and rely solely on data from the target dataset.

\textbf{Architecture}: The models with a 1x1 input window configuration were implemented as ARIMA models, a widely used time series forecasting method \citep{box_time_2015}. These models are developed to capture temporal patterns and make predictions based on historical data. The models with large input windows were implemented using the ConvLSTM architecture, structured to predict the next value in the time series based on the input window \citep{shi_convolutional_2015}. 

\subsubsection{ML Models Training Regions}\label{sec_training_regions} We also explored different training data regions and their impact on the selection of the ML models.

\textbf{Spatial Regions:} Our experiments involved three discrete training regions, each corresponding to one of the clusters identified through the Parcorr representation. Suppose the target region of the query matches the series assigned to a cluster. In that case, models trained on that specific cluster tend to have an advantage in predicting that region's accuracy. 

\textbf{Training Timestamps:}
Our experiments also took into account the temporal dimension. We considered two distinct training intervals. The first interval spanned the initial three months of the first semester of 2014, aligning with the southern hemisphere's summer season, characterized by warmer temperature patterns. In contrast, the second interval encompassed the first three months of the second semester of 2014, coinciding with the southern hemisphere's winter season, typified by cooler temperature patterns.

We trained 24 distinct models for each training timestamp, encompassing all possible combinations of input window size, training timestamps, and training regions. The input size for all models remained consistent at 10.

\subsection{Effect of the Series Representation}

StreamEnsemble establishes that the time series should be represented in a lower dimensional space. We compared two dimensionality reduction techniques, Generalized Lambda Distribution (GLD) and Parcorr, discussed in Section \ref{sec_query_partitioning}.

In the case of the Parcorr representation, we explored its sensitivity to the number of basis vectors used for dimensionality reduction. This involved varying the number of basis vectors in 2, 4, 6, and 8 vectors. By systematically adjusting this parameter, we aimed to assess the impact of different dimensional reductions on the algorithm's performance and effectiveness. 

For this experiment, the time series was fixed at 1460 data points, corresponding to an entire year of data spanning 2015. This choice of data length allows us to capture and analyze seasonal and yearly patterns in the time series. 

We employed the silhouette score to assess the series representation's quality. The silhouette score measures the clustering quality regarding the average distance between clusters and the average distance within clusters — higher scores indicate better separation. We explored different numbers of clusters, ranging from 3 to 5. Across all dimensionality reduction methods tested, it was consistently observed that the optimal number of clusters, as determined by the silhouette score, remained at 3.

Table \ref{tab_method} indicates the average silhouette values obtained after running each experiment 30 times. In the first column, the identifiers \textit{Parcorr n} indicate the parcorr dimensionality reduction (DR) approach considering $n$ basis vectors. It also displays the mean clustering time. 

\begin{figure}[!ht]
	\centering	
 \includegraphics[width=1.0\columnwidth]{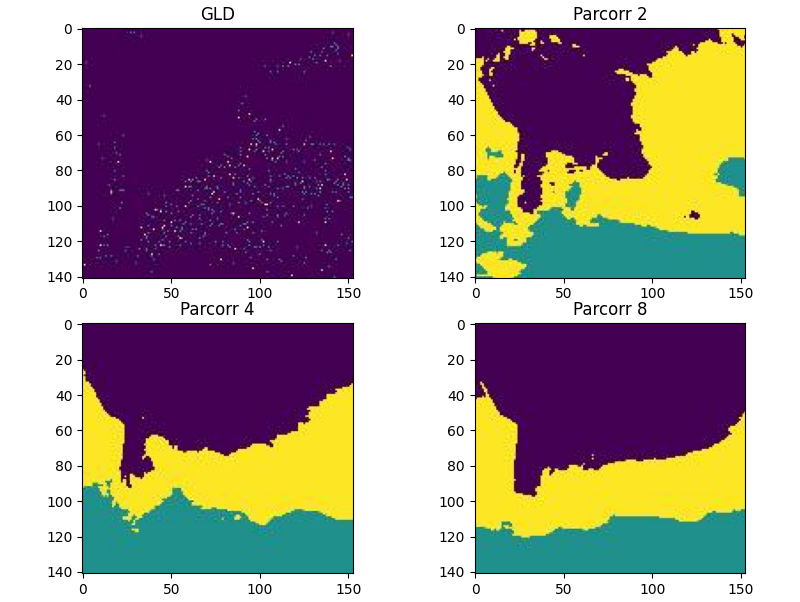} 
	\caption{Clusters identified by GLD and Parcorr with different numbers of basis vectors}
	\label{fig_parcorr_gld}
\end{figure}

Figure \ref{fig_parcorr_gld} illustrates the representation quality of GLD against Parcorr. We can see that though GLD exhibits a significantly higher silhouette score, its solution is far from optimal, as the resulting clusters become very imbalanced. This factor may indicate a sensitivity to fluctuations in one or more of the four GLD parameters that may not be directly relevant to clustering or the presence of outliers in the data. Notably, Parcorr, with just two basis vectors, proved sufficient for effective clustering, showcasing its ability to identify distinct regions. Additionally, Parcorr demonstrated an exceptionally faster processing speed, making it a more efficient choice, particularly in online stream data analysis. 

\begin{table}[!ht]
\centering
\begin{tabular}{|c|c|c|c|c|c|}
\hline
DR Method & DR Time & Clustering Time & Silhouette Score \\
\hline
GLD & 222.2560 & 25.5749 & 0.9634 \\
Parcorr2 & 1.5028 & 24.8606 & 0.5304 \\
Parcorr4 & 1.8599 & 25.2249 & 0.4997 \\
Parcorr6 & 2.4037 & 25.3640 & 0.5056 \\
Parcorr8 & 2.9887 & 25.3572 & 0.4968 \\
\hline
\end{tabular}
\caption{Results for GLD and Parcorr Dimensionality Reduction (DR) methods.}
\label{tab_method}
\end{table}

\subsection{Effect of Clustering Strategy}

The second set of experiments aimed to evaluate the time series clustering strategies (Static, Dynamic, and Stream Clustering). Each strategy was tested in isolation to assess its effectiveness in partitioning the data for subsequent analysis.

The evaluations were performed under a tumbling window model. To ensure a broader dataset coverage, each window has a fixed size of 24 data points, corresponding to six input days, without intersection between subsequent windows and across all year of 2015, resulting in 60 windows. We executed each configuration five times, resulting in 300 executions per strategy. The silhouette scores were computed after each clustering process to provide insights into the ability of each strategy to create homogeneous clusters from the input data. 

\begin{figure}[!ht]
	\centering	
 \includegraphics[width=0.9\columnwidth]{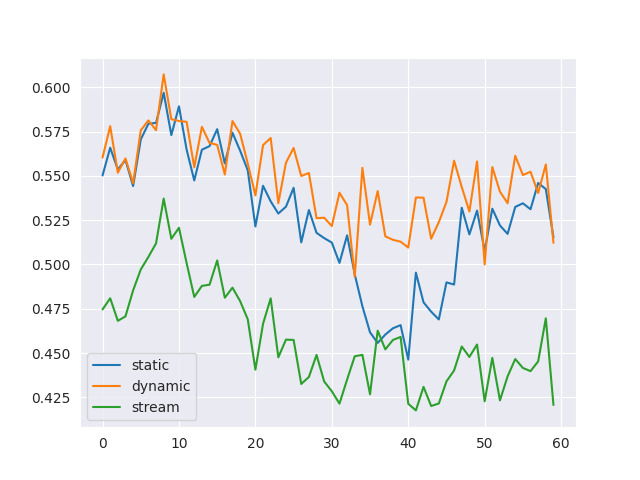} 
	\caption{Silhouette History}
	\label{fig_silhouette_through_time}
\end{figure}

Figure \ref{fig_silhouette_through_time} shows the silhouette score after each window execution. Analyzing the results, we observed a fluctuating performance for Static Clustering, decreasing during windows more representative of the winter period and increasing again in the summer. Static and Dynamic strategies performed similarly in the initial time windows, with little difference upon reclustering. However, over time, Dynamic Clustering displayed a tendency for improvement compared to Static Clustering, maintaining consistently high performance even during winter. In contrast, Stream Clustering did not exhibit a subsequent improvement in performance over time. These findings underscore the varying effectiveness of clustering strategies across different temporal contexts. Dynamic clustering stands out for its sustained high performance.

\begin{figure}[!ht]
	\centering	
 \includegraphics[width=1.0\columnwidth]{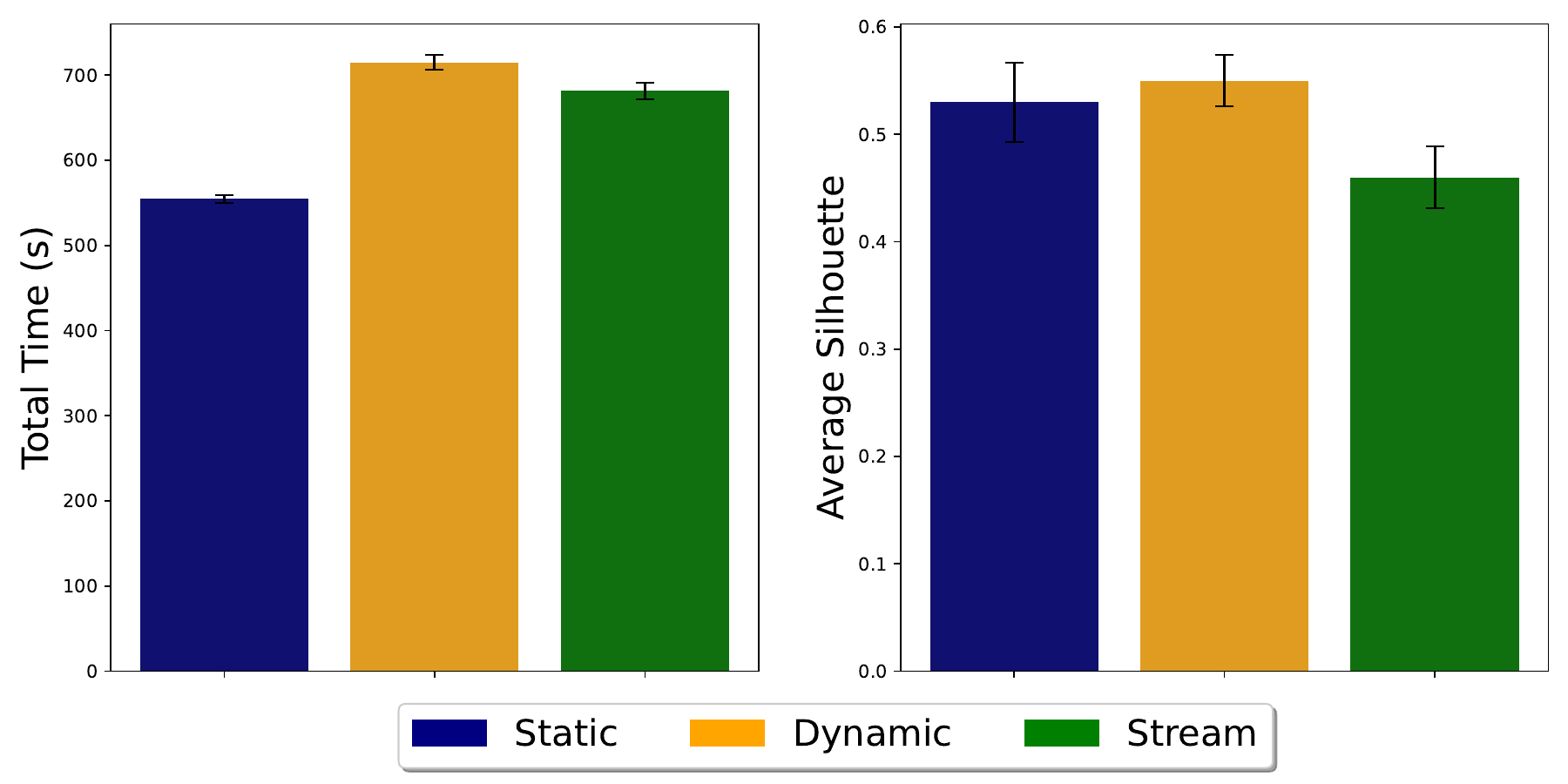} 
	\caption{Results of Clustering Strategies: Execution time and silhouette}	\label{fig_clustering-static-dynamic}
\end{figure}

These results are summarized in Figure \ref{fig_clustering-static-dynamic}. Stream clustering exhibited the least favorable results within this scenario. As expected, dynamic clustering yields the highest silhouette score, indicating superior clustering quality. However, the entire process also leads to the longest execution time. It indicates that the best strategy is a balanced approach that executes clustering periodically while avoiding reclustering every window. 

\subsection{End-to-end evaluation of the StreamEnsemble}

The final set of experiments encompassed all the method's steps, from start to finish. The purpose was to evaluate the method's overall performance, considering all its components and stages, including series representation, clustering, model selection, and execution. For this evaluation, we measured execution time and accuracy for each predictive window, assessed through the Root Mean Square Error (RMSE). 

We selected seven queries within the CFSR data to account for data variability, resulting in 7 different datasets. We considered as input the totality of data windows throughout a year-long continuous stream. Tumbling windows were employed to ensure a consistent and non-overlapping data partitioning approach. The comparative analysis considered our proposed method, integrating all 24 models described in Section \ref{sec_training_regions} over spatial queries of 20 x 20 (lat x lon). 

\begin{figure}[!ht]
	\centering	
 \includegraphics[width=1.0\columnwidth]{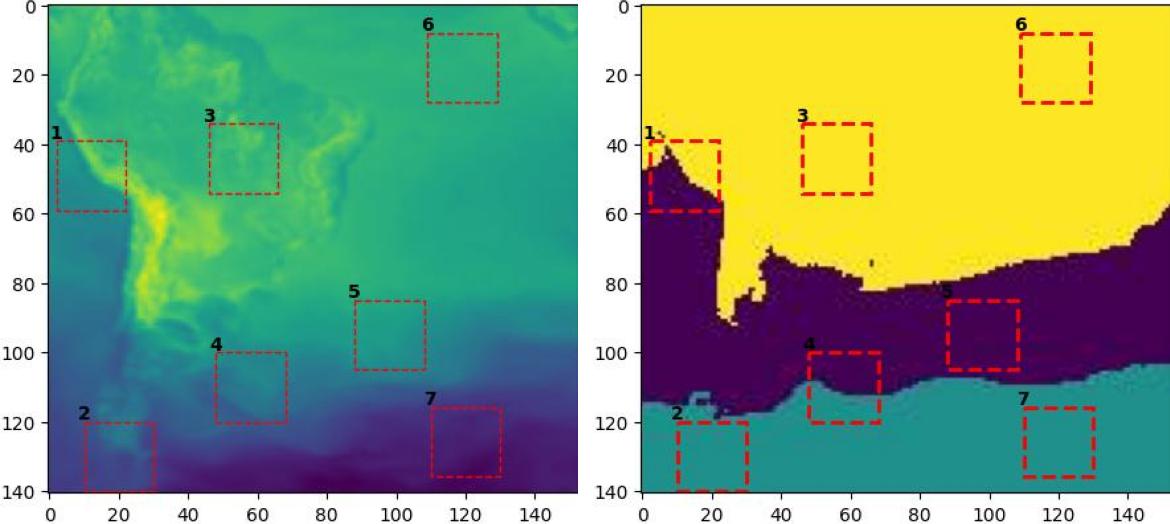} 
	\caption{Regions where each dataset was extracted from. To determine each dataset location, we performed a reference clustering for a whole year of data}	
 \label{fig_queries}
\end{figure}

Figure \ref{fig_queries} illustrates the regions where  each dataset was extracted from. To determine regions, we performed a single clustering of data over one year. We specified 5 datasets that covered a single cluster each (2, 3, 5, 6, and 7) and two datasets over the intersection of clusters (1 and 4). Our approach involved implementing static, stream, and dynamic clustering strategies over each dataset on time series with reduced dimensionality using the Parcorr method with six basis vectors. We also compared the results using the YOLO and Quadtree tiling strategies introduced in Section \ref{sec_tiling}, registering the predictive error that each strategy presented for each window. 

\begin{figure}[!ht]
\hspace*{-0.5cm} 
	\centering	
 \includegraphics[width=0.95\columnwidth]{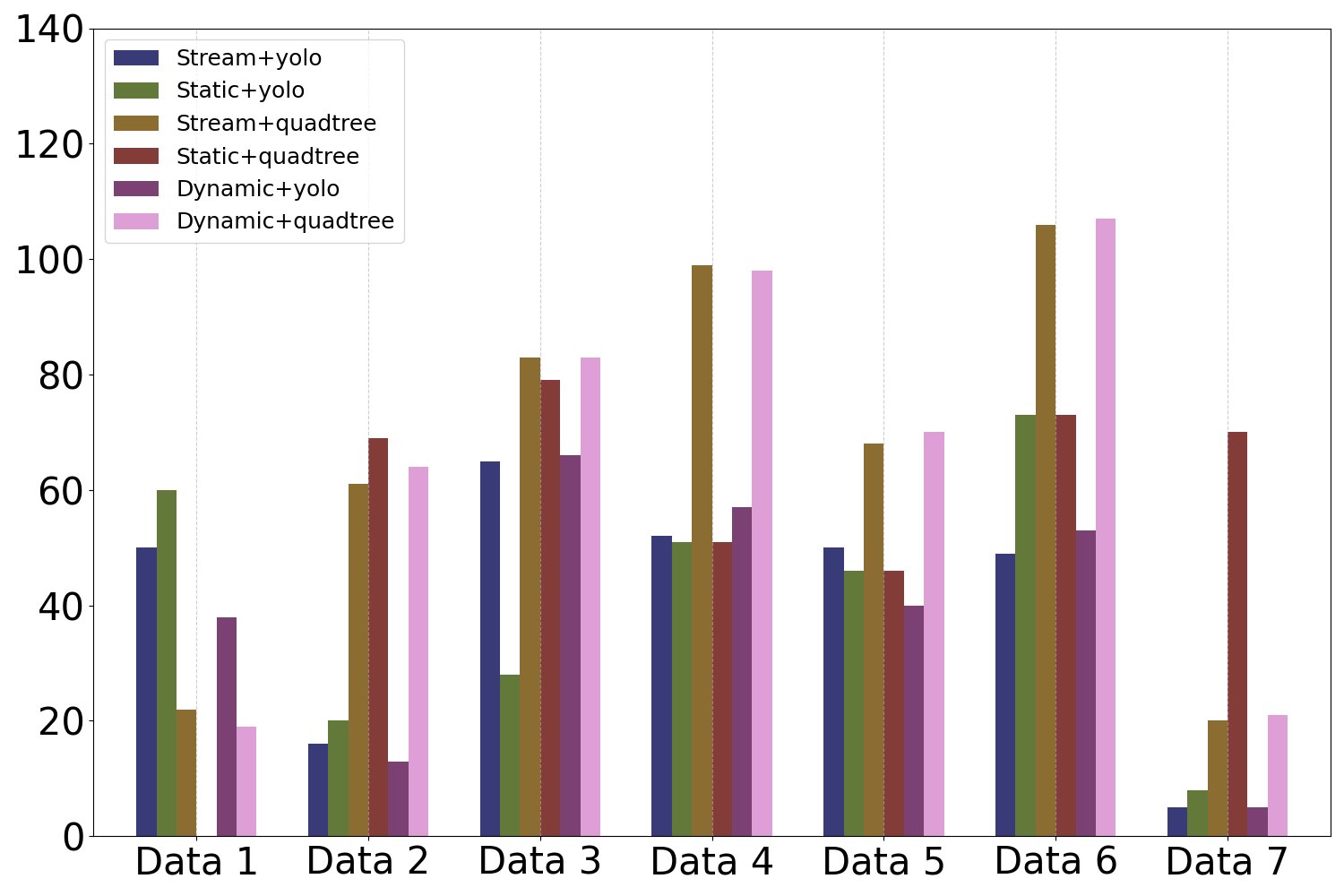} 
	\caption{Comparison between clustering and tiling strategies in different datasets. The y-axis indicates the number of windows where a strategy achieved the best results}	\label{fig_strategies}
\end{figure}

Results can be seen in Figure \ref{fig_strategies}. In it, we display the number of data windows where each strategy chose the model with the best results. When multiple strategies yielded the models with minimum error, all such strategies were considered the most successful selections. There isn't a single strategy that consistently outperforms all others across every dataset. Thus, the optimal approach depends on the characteristics of the query being processed. We also observed that strategies based on Yolo tiling tend to exhibit a higher execution time due to its bottom-up nature, leading to the identification of more tiles than quadtree and, therefore, requiring more model executions. Conversely, quadtree-based strategies demonstrated very low execution time due to fewer identified tiles. The strategy Static+quadtree took, on average 1.66 seconds to run (per window) against 10.58 for Static+yolo. 

It is worth noting that for Dataset 7, the static + quadtree strategy consistently outperformed all others. This dataset covers a region with lower temperature variability, favoring a smaller number of tiles and avoiding model switching. Nonetheless, even in this scenario, dynamic and stream-based strategies presented competitive results, with an average error difference of no more than 1.86.

We also compared the performance of our proposed method against four baseline strategies to provide a comprehensive analysis of its effectiveness. The first baseline strategy, Random Allocation, randomly selects a model to perform a prediction for each window without considering the characteristics of the data or the models. For the second strategy, Global Allocation, we trained a single model using data from multiple clusters to include variations in the data distribution. The third strategy, Average Allocation, follows a more traditional ensemble approach: it aggregates the outputs of all available models and presents the average prediction as the final result. Lastly, the ``Best of All'' strategy selects each available model to predict all time windows independently and calculate the average error for each model. Ultimately, the prediction with the lowest average error is considered the final output. These baseline strategies serve as reference points for evaluating the performance of our proposed method. 

The predictive errors of the comparative analysis are presented in Table \ref{tab_baseline}. Clearly, StreamEnsemble outperforms the Average, Random, and Global baselines, demonstrating its superiority over single-model and simple averaging-based strategies. Additionally, except for Dataset 1, in all cases, it achieved performance comparable to or better than the best possible single model, which indicates that the method effectively harnesses the strengths of multiple models to achieve competitive prediction accuracy even when no knowledge is provided of which model is optimal for each given scenario. These findings underscore the effectiveness of our method in selecting the best-suited models to improve prediction accuracy.

Notably, the Global Allocation method incurs the highest error. This suggests that the number of model parameters might be insufficient to capture the full range of data distributions within each region. Consequently, this limitation leads to elevated error rates, supporting our initial premise. The Random Allocation method exhibited the largest standard deviation in prediction error among all strategies. Since each model is chosen randomly, its performance is inconsistent and less reliable compared to other strategies.

\begin{table}[!ht]
\centering
\begin{tabular}{|c|c|c|c|c|c|}
\hline
Dataset & St Ensemble & Average & Random & Global & Best \\
\hline
1 & 4.09 & 7.91 & 12.62 & 24.61 & \textcolor{red}{3.15}\\
2 &\textcolor{red}{2.52} & 5.92 & 14.49 & 29.20 & \textcolor{red}{2.52} \\
3 &\textcolor{red}{2.25} & 12.05 & 12.83 & 18.81 & 2.28 \\
4 &\textcolor{red}{3.38} & 3.69 & 13.52 & 25.72 & 3.49 \\
5 &\textcolor{red}{2.66} & 3.48 & 10.30 & 22.79 & 2.80 \\
6 &\textcolor{red}{0.72} & 10.61 & 11.76 & 16.99 & 0.97 \\
7 &\textcolor{red}{2.71} & 8.26 & 16.57 & 31.45 & \textcolor{red}{2.71} \\
\hline
\end{tabular}
\caption{Baseline comparison of StreamEnsemble with different strategies. }
\label{tab_baseline}
\end{table}

\section{Related Work}
\label{sec_related_work}

The related work presents ML approaches for streaming data. It is organized into four sub-sections: Concept Drift Adaptation Methods, Ensemble Algorithms, Model Selection Systems, and AutoML. Each section explores methods to tackle evolving data streams using automated ML processes. These discussions offer a comprehensive view of adaptive learning techniques in response to dynamic data and predictive modeling needs.

\subsection{Concept Drift Adaptation Methods}
Concept drift, a prevalent challenge in evolving data streams, has spurred the development of various adaptation methods within ML. These methods aim to address the dynamic nature of data distributions over time, ensuring the sustained efficacy of predictive models. Techniques such as Online Random Forests \citep{saffari_-line_2009} continuously update models as new data arrives, adapting to changing concepts without retraining the entire model. Other approaches, like Page-Hinkley and ADWIN \citep{bifet_learning_2007}, monitor data streams for significant changes, signaling the occurrence of concept drift, for model adaptation or retraining \citep{baier_handling_2020}. According to \citep{lu_learning_2019}, there are three main approaches to address concept drift: simple retraining, ensemble retraining, and model adjusting.

\textbf{Simple retraining} algorithms, like Paired Learners \citep{bach_paired_2008}, use a windowing strategy, retaining recent data for retraining while considering older data for distribution change tests. Trade-offs exist when choosing window size: smaller windows reflect the latest data, but larger ones provide more training data.

\textbf{Ensemble retraining} seeks to handle concept drift by preserving and reusing a combination of previously trained models. This strategy can notably reduce the effort required to train new models, especially in recurring concept drifts. Bagging, Boosting, and Random Forests have all been extended for handling streaming data with concept drift \citep{bifet_learning_2007, lu_learning_2019}. Techniques such as Dynamic Weighted Majority (DWM) \citep{kolter_dynamic_2007} adapt to drifts using weighted voting rules, occasionally adding or removing base classifiers from the ensemble.

\textbf{Model adjusting} involves adapting models based on changing data, especially effective for localized drift. Online decision tree algorithms like VFDT \citep{domingos_mining_2000} and its extension, CVFDT \citep{hulten_mining_2001}, are designed for high-speed data streams. VFDT utilizes Hoeffding, which is bound to limit instances needed for node splitting and requires minimal storage and processing. At the same time, CVFDT employs a sliding window to handle concept drift, replacing obsolete subtrees with newer, better-performing alternatives.

These methods and their variants have offered promising strategies for managing concept drift, providing adaptive solutions to sustain model performance in the face of evolving data streams.
In contrast, our proposed work emphasizes selecting models based on the specific characteristics of the data distribution for each data window, offering a distinct approach to handling concept drift within this context, specifically for ST data.

\subsection{Ensemble Algorithms}

Ensemble learning is a term used for a category of methods that combine multiple inducers to make a decision \citep{sagi_ensemble_2018}. These methods seek to integrate several ML-based models into a unified structure so that the complimentary learned information of each inducer can contribute to increasing the performance of the overall model. The combination strategy can be either by assigning different weights to each model's Prediction or by a method that chooses the most adequate model for each given situation using some criteria. In the first case, the combination of each model's predictions can be achieved (typically) by a majority voting system for classification problems or by averaging the output of each inducer for regression problems \citep{mendes-moreira_ensemble_2012}. Ensemble learning has been appointed as one of the most effective research directions to address several challenges for increasing the performance of ML algorithms \citep{krawczyk_ensemble_2017, dong_survey_2020}. 
 
Another motivation for combining different approaches comes from the \textit{no free lunch theorem} \citep{wolpert_supervised_2002}, which states that, in a system based on certain constraints, ``over the space of all possible problems, every optimization technique will perform as well as every other one on average'' \citep{lones_sean_2011}. The statement indicates that building a single classifier simultaneously adequate for all problems is not feasible, as each classifier will have its domain of expertise.
 
Some widely used ensemble classification methods include bagging, AdaBoost, random forest, random subspace, and gradient boosting \citep{sagi_ensemble_2018, dong_survey_2020}. These methods use a single learning algorithm and generate different classifiers by manipulating the training set input or the algorithm's output \citep{dzeroski_is_2004}. They also tend to be more computationally efficient when the execution cost of the participating inducers is low \citep{sagi_ensemble_2018}. 

Another approach involves applying heterogeneous learning algorithms with distinct model representations to a single dataset. In this context, more intricate methods for combining classifiers are commonly employed. Model Stacking \citep{pavlyshenko_using_2018, wolpert_stacked_1992} is a frequently used method for learning how to combine classifiers and the existing ensemble of classifiers. Voting or averaging the predictions is a benchmark for comparing the performance of the learned combiners to that of the ensemble \citep{dzeroski_is_2004}.

Several adaptations to some of these methods have been proposed for dealing with stream data \citep{krawczyk_ensemble_2017}. However, many focus on building new classifiers upon detecting time series distribution changes. Most do not specifically address the characteristics of ST data streams. 

Ensemble-based strategies like Dynamic Weighted Majority \citep{kolter_dynamic_2007} and Learn++ \citep{polikar_learn_2001} as well as its variations (e.g., \citep{ditzler_incremental_2010, elwell_incremental_2011}) leverage multiple models, dynamically weighting their predictions or adjusting model weights based on the evolving data patterns. Most of these existing ensemble algorithms are homogeneous and derived from a single type of learning model.

\citet{jiang_spatial_2019} proposed an ensemble learning approach designed to handle spatial data where regions demonstrate heterogeneous data distributions. It also presents a strategy to partition the dataset geographically based on the characteristics of each region. However, it does not assume pre-trained models for the ensemble construction; instead, it trains new basic models for each identified zone within the data. Additionally, it does not handle ST data and is designed exclusively for classification problems.

\citet{luong_heterogeneous_2021} addressed a similar scenario. They propose a method denominated Heterogeneous Ensemble Selection (HEES), which focuses on the composition of a model ensemble of several potentially heterogeneous algorithms to handle evolving data streams. However, while our approach directly incorporates the time series characteristics when composing the ensemble, their method predominantly relies on model accuracy for ensemble selection. Moreover, our approach is designed specifically for the ST series. Lastly, HEES is specifically tailored for classification tasks and does not extend its application to regression problems, diverging from the scope of our proposed approach.

\citet{pereira_djensemble_2021} provide a similar scenario as the one we propose. However, its core focus remains on handling static ST datasets. It does not possess the capability to manage evolving stream data effectively.

\subsection{Model Selection Management Sytems}

Model selection management systems (MSMS) such as DLHub \citep{chard_dlhub_2019} and Cerebro \citep{nakandala_cerebro_2020} are comprehensive frameworks designed to streamline the process of choosing, evaluating, and deploying ML models within diverse applications \citep{schelter_challenges_2018}. These systems serve as centralized platforms that facilitate efficient handling of various stages in the model selection pipeline, encompassing functionalities for data preprocessing, feature engineering, model training, validation, and hyperparameter tuning \citep{kumar_model_2016}.
Moreover, they frequently integrate version control and tracking mechanisms to monitor model performance over time, aiding in the decision-making process for model updates or replacements as new data becomes available or as algorithms evolve. 

MSMS systems focus on providing a centralized infrastructure for managing the entire process of selecting ML models across various stages of development. In contrast, our method assumes pre-trained ML models that must be combined to solve ST queries over continuous data streams, requiring real-time or near-real-time model selection. The challenges of our proposed algorithm involve the unique characteristics of streaming data, such as concept drift and adaptive learning, and selecting among a set of pre-trained ST models rather than providing a broader framework for managing all aspects of model training and selection across various domains.

\subsection{AutoML}

AutoML (Automated ML) frameworks have gained significant attention recently, offering automated solutions for model selection, hyperparameter tuning, and ensemble construction \citep{hutter_automated_2019}. These frameworks aim to streamline the ML pipeline by automating various stages, including data preprocessing, feature engineering, model selection, and optimization. While AutoML approaches provide valuable automation and optimization capabilities, they often focus on handling tabular or static datasets with predefined features, lacking specialized adaptability to complex ST data streams.

StreamEnsemble, as proposed, can serve as a valuable contribution within the scope of Model Serving Systems and AutoML strategies. Model selection, an important point in AutoML \citep{hutter_automated_2019}, is integral to several cutting-edge predictor serving systems like Clipper \citep{crankshaw_clipper_2017} and Rafiki \citep{wang_rafiki_2018}. These systems prioritize model selection to efficiently serve predictions based on the specific requirements of the target data.

Our proposed StreamEnsemble operates by considering a cost model that statically outlines a prediction ensemble plan reliant on estimated accuracy and prediction time. This methodology aligns with the principles of model selection. It could enhance existing AutoML frameworks by extending their adaptability to dynamic ST datasets. StreamEnsemble can function as a model selection solution when integrated into systems offering auto-regressive ST Prediction (STP) services. While AutoML frameworks excel in automating model selection and hyperparameter tuning, StreamEnsemble extends beyond by addressing the nuances of spatially distributed and evolving temporal data, ensuring adaptability to changing data distributions and spatial complexities in real-time analysis scenarios.

\section{Conclusion}
\label{sec_conclusion}

In this paper, we proposed StreamEnsemble, a novel approach to predictive queries over ST data that performs a dynamic ensemble selection of ML models based on the characteristics of data distributions. StreamEnsemble addresses the challenge of efficiently allocating models to evolving data patterns in a streaming environment. The algorithm employs a three-step approach: estimating models' generalization errors, clustering time series representations, and defining ensembles tailored to specific data partitions. It dynamically adapts to evolving data distributions, effectively selecting and allocating ML models to answer continuous queries in real time. Furthermore, it incorporates techniques like dynamic clustering and hierarchical tiling strategies, ensuring adaptability and precision in handling irregular and noisy datasets. By transforming the irregular clustering into regular structures required by ST models, our approach enhances the accuracy of predictive queries in ST streaming environments. 

There are promising directions for future exploration and enhancement of the proposed algorithm. As a potential direction for future research, we propose exploring an active ensemble method that dynamically compares incoming errors with a predefined reference threshold rather than continually rebuilding the ensemble. We also mention the need to investigate further the impact of the clustering and tiling strategies and the dimensionality reduction function. Lastly, another area of interest lies in exploring the application of this algorithm in specific domains to evaluate its effectiveness in real-world scenarios and identify domain-specific optimizations. 

\begin{acks}

This work has been partially supported by CAPES, CNPq, and FAPERJ in Brazil and by the HPDaSc Inria associated team.

\end{acks}

\balance
\bibliographystyle{ACM-Reference-Format}
\bibliography{references}


\begin{thebibliography}{61}


\ifx \showCODEN    \undefined \def \showCODEN     #1{\unskip}     \fi
\ifx \showDOI      \undefined \def \showDOI       #1{#1}\fi
\ifx \showISBNx    \undefined \def \showISBNx     #1{\unskip}     \fi
\ifx \showISBNxiii \undefined \def \showISBNxiii  #1{\unskip}     \fi
\ifx \showISSN     \undefined \def \showISSN      #1{\unskip}     \fi
\ifx \showLCCN     \undefined \def \showLCCN      #1{\unskip}     \fi
\ifx \shownote     \undefined \def \shownote      #1{#1}          \fi
\ifx \showarticletitle \undefined \def \showarticletitle #1{#1}   \fi
\ifx \showURL      \undefined \def \showURL       {\relax}        \fi
\providecommand\bibfield[2]{#2}
\providecommand\bibinfo[2]{#2}
\providecommand\natexlab[1]{#1}
\providecommand\showeprint[2][]{arXiv:#2}

\bibitem[\protect\citeauthoryear{Afyouni, Khan, and Aghbari}{Afyouni et~al\mbox{.}}{2022}]%
        {afyouni_deep-eware_2022}
\bibfield{author}{\bibinfo{person}{Imad Afyouni}, \bibinfo{person}{Aamir Khan}, {and} \bibinfo{person}{Zaher~Al Aghbari}.} \bibinfo{year}{2022}\natexlab{}.
\newblock \showarticletitle{Deep-{Eware}: spatio-temporal social event detection using a hybrid learning model}.
\newblock \bibinfo{journal}{\emph{Journal of Big Data}} \bibinfo{volume}{9}, \bibinfo{number}{1} (\bibinfo{year}{2022}).
\newblock
\urldef\tempurl%
\url{https://doi.org/10.1186/s40537-022-00636-w}
\showDOI{\tempurl}


\bibitem[\protect\citeauthoryear{Aghabozorgi, {Seyed Shirkhorshidi}, and {Ying Wah}}{Aghabozorgi et~al\mbox{.}}{2015}]%
        {aghabozorgi_time-series_2015}
\bibfield{author}{\bibinfo{person}{Saeed Aghabozorgi}, \bibinfo{person}{Ali {Seyed Shirkhorshidi}}, {and} \bibinfo{person}{Teh {Ying Wah}}.} \bibinfo{year}{2015}\natexlab{}.
\newblock \showarticletitle{Time-series clustering - {A} decade review}.
\newblock \bibinfo{journal}{\emph{Information Systems}}  \bibinfo{volume}{53} (\bibinfo{year}{2015}), \bibinfo{pages}{16 -- 38}.
\newblock
\urldef\tempurl%
\url{https://doi.org/10.1016/j.is.2015.04.007}
\showDOI{\tempurl}


\bibitem[\protect\citeauthoryear{Atluri, Karpatne, and Kumar}{Atluri et~al\mbox{.}}{2018}]%
        {atluri_spatio-temporal_2018}
\bibfield{author}{\bibinfo{person}{Gowtham Atluri}, \bibinfo{person}{Anuj Karpatne}, {and} \bibinfo{person}{Vipin Kumar}.} \bibinfo{year}{2018}\natexlab{}.
\newblock \showarticletitle{Spatio-temporal data mining: {A} survey of problems and methods}.
\newblock \bibinfo{journal}{\emph{Comput. Surveys}} \bibinfo{volume}{51}, \bibinfo{number}{4} (\bibinfo{year}{2018}).
\newblock
\urldef\tempurl%
\url{https://doi.org/10.1145/3161602}
\showDOI{\tempurl}


\bibitem[\protect\citeauthoryear{Bach and Maloof}{Bach and Maloof}{2008}]%
        {bach_paired_2008}
\bibfield{author}{\bibinfo{person}{Stephen~H. Bach} {and} \bibinfo{person}{Marcus~A. Maloof}.} \bibinfo{year}{2008}\natexlab{}.
\newblock \showarticletitle{Paired {Learners} for {Concept} {Drift}}. In \bibinfo{booktitle}{\emph{2008 {Eighth} {IEEE} {International} {Conference} on {Data} {Mining}}}. \bibinfo{publisher}{IEEE}, \bibinfo{pages}{23--32}.
\newblock
\showISBNx{978-0-7695-3502-9}
\urldef\tempurl%
\url{https://doi.org/10.1109/ICDM.2008.119}
\showDOI{\tempurl}


\bibitem[\protect\citeauthoryear{Baier, Reimold, and Kuhl}{Baier et~al\mbox{.}}{2020}]%
        {baier_handling_2020}
\bibfield{author}{\bibinfo{person}{Lucas Baier}, \bibinfo{person}{Josua Reimold}, {and} \bibinfo{person}{Niklas Kuhl}.} \bibinfo{year}{2020}\natexlab{}.
\newblock \showarticletitle{Handling {Concept} {Drift} for {Predictions} in {Business} {Process} {Mining}}. In \bibinfo{booktitle}{\emph{Proceedings - 2020 {IEEE} 22nd {Conference} on {Business} {Informatics}, {CBI} 2020}}, \bibfield{editor}{\bibinfo{person}{Aier S}, \bibinfo{person}{Gordijn J}, \bibinfo{person}{Proper H.A}, {and} \bibinfo{person}{Verelst J}} (Eds.), Vol.~\bibinfo{volume}{1}. \bibinfo{publisher}{IEEE}, \bibinfo{pages}{76 -- 83}.
\newblock
\showISBNx{978-1-72819-926-9}
\urldef\tempurl%
\url{https://doi.org/10.1109/CBI49978.2020.00016}
\showDOI{\tempurl}


\bibitem[\protect\citeauthoryear{Bellman and Kalaba}{Bellman and Kalaba}{1958}]%
        {bellman_adaptive_1958}
\bibfield{author}{\bibinfo{person}{Richard Bellman} {and} \bibinfo{person}{Robert Kalaba}.} \bibinfo{year}{1958}\natexlab{}.
\newblock \showarticletitle{On adaptive control processes}.
\newblock \bibinfo{journal}{\emph{IRE Transactions on Automatic Control}} \bibinfo{volume}{4}, \bibinfo{number}{2} (\bibinfo{year}{1958}), \bibinfo{pages}{1 -- 9}.
\newblock
\urldef\tempurl%
\url{https://doi.org/10.1109/TAC.1959.1104847}
\showDOI{\tempurl}


\bibitem[\protect\citeauthoryear{Bergevin}{Bergevin}{1993}]%
        {bergevin_analysis_1993}
\bibfield{author}{\bibinfo{person}{Robert~J. Bergevin}.} \bibinfo{year}{1993}\natexlab{}.
\newblock \bibinfo{booktitle}{\emph{An {Analysis} of the {Generalized} {Lambda} {Distribution}}}.
\newblock \bibinfo{type}{{PhD} {Thesis}}. \bibinfo{institution}{Air Force Institute of Technology}.
\newblock


\bibitem[\protect\citeauthoryear{Bifet and Gavald{\a`a}}{Bifet and Gavald{\a`a}}{2007}]%
        {bifet_learning_2007}
\bibfield{author}{\bibinfo{person}{Albert Bifet} {and} \bibinfo{person}{Ricard Gavald{\a`a}}.} \bibinfo{year}{2007}\natexlab{}.
\newblock \showarticletitle{Learning from time-changing data with adaptive windowing}. In \bibinfo{booktitle}{\emph{Proceedings of the 7th {SIAM} {International} {Conference} on {Data} {Mining}}}. \bibinfo{publisher}{Society for Industrial and Applied Mathematics Publications}, \bibinfo{pages}{443 -- 448}.
\newblock
\showISBNx{978-0-89871-630-6}
\urldef\tempurl%
\url{https://doi.org/10.1137/1.9781611972771.42}
\showDOI{\tempurl}


\bibitem[\protect\citeauthoryear{Bonnet, Evsukoff, and Rodriguez}{Bonnet et~al\mbox{.}}{2020}]%
        {bonnet_precipitation_2020}
\bibfield{author}{\bibinfo{person}{Suzanna~Maria Bonnet}, \bibinfo{person}{Alexandre Evsukoff}, {and} \bibinfo{person}{Carlos Augusto~Morales Rodriguez}.} \bibinfo{year}{2020}\natexlab{}.
\newblock \showarticletitle{Precipitation nowcasting with weather radar images and deep learning in são paulo, brasil}.
\newblock \bibinfo{journal}{\emph{Atmosphere}} \bibinfo{volume}{11}, \bibinfo{number}{11} (\bibinfo{year}{2020}).
\newblock
\urldef\tempurl%
\url{https://doi.org/10.3390/atmos11111157}
\showDOI{\tempurl}


\bibitem[\protect\citeauthoryear{Box, Jenkins, Reinsel, and Ljung}{Box et~al\mbox{.}}{2015}]%
        {box_time_2015}
\bibfield{author}{\bibinfo{person}{George E.~P. Box}, \bibinfo{person}{Gwilym~M. Jenkins}, \bibinfo{person}{Gregory~C. Reinsel}, {and} \bibinfo{person}{Greta~M. Ljung}.} \bibinfo{year}{2015}\natexlab{}.
\newblock \bibinfo{booktitle}{\emph{Time {Series} {Analysis}: {Forecasting} and {Control}}}.
\newblock \bibinfo{publisher}{John Wiley \& Sons}.
\newblock
\showISBNx{978-1-118-67492-5}


\bibitem[\protect\citeauthoryear{Chard, Li, Chard, Ward, Babuji, Woodard, Tuecke, Blaiszik, Franklin, and Foster}{Chard et~al\mbox{.}}{2019}]%
        {chard_dlhub_2019}
\bibfield{author}{\bibinfo{person}{Ryan Chard}, \bibinfo{person}{Zhuozhao Li}, \bibinfo{person}{Kyle Chard}, \bibinfo{person}{Logan Ward}, \bibinfo{person}{Yadu Babuji}, \bibinfo{person}{Anna Woodard}, \bibinfo{person}{Steven Tuecke}, \bibinfo{person}{Ben Blaiszik}, \bibinfo{person}{Michael~J. Franklin}, {and} \bibinfo{person}{Ian Foster}.} \bibinfo{year}{2019}\natexlab{}.
\newblock \showarticletitle{{DLHub}: {Model} and data serving for science}. In \bibinfo{booktitle}{\emph{Proceedings - 2019 {IEEE} 33rd {International} {Parallel} and {Distributed} {Processing} {Symposium}, {IPDPS} 2019}}. \bibinfo{publisher}{IEEE}, \bibinfo{pages}{283 -- 292}.
\newblock
\showISBNx{978-1-72811-246-6}
\urldef\tempurl%
\url{https://doi.org/10.1109/IPDPS.2019.00038}
\showDOI{\tempurl}


\bibitem[\protect\citeauthoryear{Chattopadhyay, Hassanzadeh, and Pasha}{Chattopadhyay et~al\mbox{.}}{2020}]%
        {chattopadhyay_predicting_2020}
\bibfield{author}{\bibinfo{person}{Ashesh Chattopadhyay}, \bibinfo{person}{Pedram Hassanzadeh}, {and} \bibinfo{person}{Saba Pasha}.} \bibinfo{year}{2020}\natexlab{}.
\newblock \showarticletitle{Predicting clustered weather patterns: {A} test case for applications of convolutional neural networks to spatio-temporal climate data}.
\newblock \bibinfo{journal}{\emph{Scientific Reports}} \bibinfo{volume}{10}, \bibinfo{number}{1} (\bibinfo{year}{2020}).
\newblock
\urldef\tempurl%
\url{https://doi.org/10.1038/s41598-020-57897-9}
\showDOI{\tempurl}


\bibitem[\protect\citeauthoryear{Crankshaw, Wang, Zhou, Franklin, Gonzalez, and Stoica}{Crankshaw et~al\mbox{.}}{2017}]%
        {crankshaw_clipper_2017}
\bibfield{author}{\bibinfo{person}{Daniel Crankshaw}, \bibinfo{person}{Xin Wang}, \bibinfo{person}{Giulio Zhou}, \bibinfo{person}{Michael~J. Franklin}, \bibinfo{person}{Joseph~E. Gonzalez}, {and} \bibinfo{person}{Ion Stoica}.} \bibinfo{year}{2017}\natexlab{}.
\newblock \showarticletitle{Clipper: {A} low-latency online prediction serving system}. In \bibinfo{booktitle}{\emph{Proceedings of the 14th {USENIX} {Symposium} on {Networked} {Systems} {Design} and {Implementation}, {NSDI} 2017}}. \bibinfo{pages}{613 -- 627}.
\newblock


\bibitem[\protect\citeauthoryear{Das and Ghosh}{Das and Ghosh}{2015}]%
        {das_probabilistic_2015}
\bibfield{author}{\bibinfo{person}{Monidipa Das} {and} \bibinfo{person}{Soumya~K. Ghosh}.} \bibinfo{year}{2015}\natexlab{}.
\newblock \showarticletitle{A probabilistic approach for weather forecast using spatio-temporal inter-relationships among climate variables}. In \bibinfo{booktitle}{\emph{9th {International} {Conference} on {Industrial} and {Information} {Systems}, {ICIIS} 2014}}, \bibfield{editor}{\bibinfo{person}{Arya K.V} {and} \bibinfo{person}{Kumar S}} (Eds.). \bibinfo{publisher}{IEEE}.
\newblock
\showISBNx{978-1-4799-6499-4}
\urldef\tempurl%
\url{https://doi.org/10.1109/ICIINFS.2014.7036528}
\showDOI{\tempurl}


\bibitem[\protect\citeauthoryear{Ditzler, Muhlbaier, and Polikar}{Ditzler et~al\mbox{.}}{2010}]%
        {ditzler_incremental_2010}
\bibfield{author}{\bibinfo{person}{Gregory Ditzler}, \bibinfo{person}{Michael~D. Muhlbaier}, {and} \bibinfo{person}{Robi Polikar}.} \bibinfo{year}{2010}\natexlab{}.
\newblock \showarticletitle{Incremental learning of new classes in unbalanced datasets: {Learn} ++.{UDNC}}. In \bibinfo{booktitle}{\emph{Lecture {Notes} in {Computer} {Science} (including subseries {Lecture} {Notes} in {Artificial} {Intelligence} and {Lecture} {Notes} in {Bioinformatics})}}, Vol.~\bibinfo{volume}{5997 LNCS}. \bibinfo{publisher}{Springer}, \bibinfo{pages}{33 -- 42}.
\newblock
\urldef\tempurl%
\url{https://doi.org/10.1007/978-3-642-12127-2_4}
\showDOI{\tempurl}


\bibitem[\protect\citeauthoryear{Domingos and Hulten}{Domingos and Hulten}{2000}]%
        {domingos_mining_2000}
\bibfield{author}{\bibinfo{person}{Pedro Domingos} {and} \bibinfo{person}{Geoff Hulten}.} \bibinfo{year}{2000}\natexlab{}.
\newblock \showarticletitle{Mining high-speed data streams}. In \bibinfo{booktitle}{\emph{Proceeding of the {Sixth} {ACM} {SIGKDD} {International} {Conference} on {Knowledge} {Discovery} and {Data} {Mining}}}, \bibfield{editor}{\bibinfo{person}{Ramakrishnan R}, \bibinfo{person}{Stolfo S}, \bibinfo{person}{Bayardo R}, {and} \bibinfo{person}{Parsa I}} (Eds.). \bibinfo{publisher}{Association for Computing Machinery (ACM)}, \bibinfo{pages}{71 -- 80}.
\newblock
\showISBNx{1-58113-233-6 978-1-58113-233-5}
\urldef\tempurl%
\url{https://doi.org/10.1145/347090.347107}
\showDOI{\tempurl}


\bibitem[\protect\citeauthoryear{Dong, Yu, Cao, Shi, and Ma}{Dong et~al\mbox{.}}{2020}]%
        {dong_survey_2020}
\bibfield{author}{\bibinfo{person}{Xibin Dong}, \bibinfo{person}{Zhiwen Yu}, \bibinfo{person}{Wenming Cao}, \bibinfo{person}{Yifan Shi}, {and} \bibinfo{person}{Qianli Ma}.} \bibinfo{year}{2020}\natexlab{}.
\newblock \showarticletitle{A survey on ensemble learning}.
\newblock \bibinfo{journal}{\emph{Frontiers of Computer Science}} \bibinfo{volume}{14}, \bibinfo{number}{2} (\bibinfo{year}{2020}), \bibinfo{pages}{241 -- 258}.
\newblock
\urldef\tempurl%
\url{https://doi.org/10.1007/s11704-019-8208-z}
\showDOI{\tempurl}


\bibitem[\protect\citeauthoryear{D{\v z}eroski and {\v Z}enko}{D{\v z}eroski and {\v Z}enko}{2004}]%
        {dzeroski_is_2004}
\bibfield{author}{\bibinfo{person}{Saso D{\v z}eroski} {and} \bibinfo{person}{Bernard {\v Z}enko}.} \bibinfo{year}{2004}\natexlab{}.
\newblock \showarticletitle{Is combining classifiers with stacking better than selecting the best one?}
\newblock \bibinfo{journal}{\emph{Machine Learning}} \bibinfo{volume}{54}, \bibinfo{number}{3} (\bibinfo{year}{2004}), \bibinfo{pages}{255 -- 273}.
\newblock
\urldef\tempurl%
\url{https://doi.org/10.1023/B:MACH.0000015881.36452.6e}
\showDOI{\tempurl}


\bibitem[\protect\citeauthoryear{Eklund, Nichols, and Knutsson}{Eklund et~al\mbox{.}}{2016}]%
        {eklund_cluster_2016}
\bibfield{author}{\bibinfo{person}{Anders Eklund}, \bibinfo{person}{Thomas~E. Nichols}, {and} \bibinfo{person}{Hans Knutsson}.} \bibinfo{year}{2016}\natexlab{}.
\newblock \showarticletitle{Cluster failure: {Why} {fMRI} inferences for spatial extent have inflated false-positive rates}.
\newblock \bibinfo{journal}{\emph{Proceedings of the National Academy of Sciences of the United States of America}} \bibinfo{volume}{113}, \bibinfo{number}{28} (\bibinfo{year}{2016}), \bibinfo{pages}{7900 -- 7905}.
\newblock
\urldef\tempurl%
\url{https://doi.org/10.1073/pnas.1602413113}
\showDOI{\tempurl}


\bibitem[\protect\citeauthoryear{Elwell and Polikar}{Elwell and Polikar}{2011}]%
        {elwell_incremental_2011}
\bibfield{author}{\bibinfo{person}{Ryan Elwell} {and} \bibinfo{person}{Robi Polikar}.} \bibinfo{year}{2011}\natexlab{}.
\newblock \showarticletitle{Incremental learning of concept drift in nonstationary environments}.
\newblock \bibinfo{journal}{\emph{IEEE Transactions on Neural Networks}} \bibinfo{volume}{22}, \bibinfo{number}{10} (\bibinfo{year}{2011}), \bibinfo{pages}{1517 -- 1531}.
\newblock
\urldef\tempurl%
\url{https://doi.org/10.1109/TNN.2011.2160459}
\showDOI{\tempurl}


\bibitem[\protect\citeauthoryear{Fournier, Rupin, Bigerelle, Najjar, Iost, and Wilcox}{Fournier et~al\mbox{.}}{2007}]%
        {fournier_estimating_2007}
\bibfield{author}{\bibinfo{person}{B. Fournier}, \bibinfo{person}{N. Rupin}, \bibinfo{person}{M. Bigerelle}, \bibinfo{person}{D. Najjar}, \bibinfo{person}{A. Iost}, {and} \bibinfo{person}{R. Wilcox}.} \bibinfo{year}{2007}\natexlab{}.
\newblock \showarticletitle{Estimating the parameters of a generalized lambda distribution}.
\newblock \bibinfo{journal}{\emph{Computational Statistics and Data Analysis}} \bibinfo{volume}{51}, \bibinfo{number}{6} (\bibinfo{year}{2007}), \bibinfo{pages}{2813 -- 2835}.
\newblock
\urldef\tempurl%
\url{https://doi.org/10.1016/j.csda.2006.09.043}
\showDOI{\tempurl}


\bibitem[\protect\citeauthoryear{Gomes, Barddal, {{Enembreck}}, {{Fabricio}}, and Bifet}{Gomes et~al\mbox{.}}{2017}]%
        {gomes_survey_2017}
\bibfield{author}{\bibinfo{person}{Heitor~Murilo Gomes}, \bibinfo{person}{Jean~Paul Barddal}, \bibinfo{person}{{{Enembreck}}}, \bibinfo{person}{{{Fabricio}}}, {and} \bibinfo{person}{Albert Bifet}.} \bibinfo{year}{2017}\natexlab{}.
\newblock \showarticletitle{A survey on ensemble learning for data stream classification}.
\newblock \bibinfo{journal}{\emph{Comput. Surveys}} \bibinfo{volume}{50}, \bibinfo{number}{2} (\bibinfo{year}{2017}).
\newblock
\urldef\tempurl%
\url{https://doi.org/10.1145/3054925}
\showDOI{\tempurl}


\bibitem[\protect\citeauthoryear{Hamdi, Shaban, Erradi, Mohamed, Rumi, and Salim}{Hamdi et~al\mbox{.}}{2022}]%
        {hamdi_spatiotemporal_2022}
\bibfield{author}{\bibinfo{person}{Ali Hamdi}, \bibinfo{person}{Khaled Shaban}, \bibinfo{person}{Abdelkarim Erradi}, \bibinfo{person}{Amr Mohamed}, \bibinfo{person}{Shakila~Khan Rumi}, {and} \bibinfo{person}{Flora~D. Salim}.} \bibinfo{year}{2022}\natexlab{}.
\newblock \showarticletitle{Spatiotemporal data mining: a survey on challenges and open problems}.
\newblock \bibinfo{journal}{\emph{Artificial Intelligence Review}} \bibinfo{volume}{55}, \bibinfo{number}{2} (\bibinfo{year}{2022}), \bibinfo{pages}{1441 -- 1488}.
\newblock
\urldef\tempurl%
\url{https://doi.org/10.1007/s10462-021-09994-y}
\showDOI{\tempurl}


\bibitem[\protect\citeauthoryear{Huang, Song, Hong, and Xie}{Huang et~al\mbox{.}}{2014}]%
        {huang_deep_2014}
\bibfield{author}{\bibinfo{person}{Wenhao Huang}, \bibinfo{person}{Guojie Song}, \bibinfo{person}{Haikun Hong}, {and} \bibinfo{person}{Kunqing Xie}.} \bibinfo{year}{2014}\natexlab{}.
\newblock \showarticletitle{Deep architecture for traffic flow prediction: {Deep} belief networks with multitask learning}.
\newblock \bibinfo{journal}{\emph{IEEE Transactions on Intelligent Transportation Systems}} \bibinfo{volume}{15}, \bibinfo{number}{5} (\bibinfo{year}{2014}), \bibinfo{pages}{2191 -- 2201}.
\newblock
\urldef\tempurl%
\url{https://doi.org/10.1109/TITS.2014.2311123}
\showDOI{\tempurl}


\bibitem[\protect\citeauthoryear{Hulten, Spencer, and Domingos}{Hulten et~al\mbox{.}}{2001}]%
        {hulten_mining_2001}
\bibfield{author}{\bibinfo{person}{Geoff Hulten}, \bibinfo{person}{Laurie Spencer}, {and} \bibinfo{person}{Pedro Domingos}.} \bibinfo{year}{2001}\natexlab{}.
\newblock \showarticletitle{Mining time-changing data streams}. In \bibinfo{booktitle}{\emph{Proceedings of the {Seventh} {ACM} {SIGKDD} {International} {Conference} on {Knowledge} {Discovery} and {Data} {Mining}}}, \bibfield{editor}{\bibinfo{person}{Provost F}, \bibinfo{person}{Srikant R}, \bibinfo{person}{Schkolnick M}, {and} \bibinfo{person}{Lee D}} (Eds.). \bibinfo{publisher}{Association for Computing Machinery (ACM)}, \bibinfo{pages}{97 -- 106}.
\newblock
\showISBNx{1-58113-391-X 978-1-58113-391-2}
\urldef\tempurl%
\url{https://doi.org/10.1145/502512.502529}
\showDOI{\tempurl}


\bibitem[\protect\citeauthoryear{Hutter, Kotthoff, and Vanschoren}{Hutter et~al\mbox{.}}{2019}]%
        {hutter_automated_2019}
\bibfield{author}{\bibinfo{person}{Frank Hutter}, \bibinfo{person}{Lars Kotthoff}, {and} \bibinfo{person}{Joaquin Vanschoren}.} \bibinfo{year}{2019}\natexlab{}.
\newblock \bibinfo{booktitle}{\emph{Automated {Machine} {Learning}: {Methods}, {Systems}, {Challenges}}}.
\newblock \bibinfo{publisher}{Springer International Publishing}.
\newblock
\showISBNx{978-3-030-05317-8}


\bibitem[\protect\citeauthoryear{Jiang, Sainju, Li, Shekhar, and Knight}{Jiang et~al\mbox{.}}{2019}]%
        {jiang_spatial_2019}
\bibfield{author}{\bibinfo{person}{Zhe Jiang}, \bibinfo{person}{Arpan~Man Sainju}, \bibinfo{person}{Yan Li}, \bibinfo{person}{Shashi Shekhar}, {and} \bibinfo{person}{Joseph Knight}.} \bibinfo{year}{2019}\natexlab{}.
\newblock \showarticletitle{Spatial ensemble learning for heterogeneous geographic data with class ambiguity}.
\newblock \bibinfo{journal}{\emph{ACM Transactions on Intelligent Systems and Technology}} \bibinfo{volume}{10}, \bibinfo{number}{4} (\bibinfo{year}{2019}).
\newblock
\urldef\tempurl%
\url{https://doi.org/10.1145/3337798}
\showDOI{\tempurl}


\bibitem[\protect\citeauthoryear{Kleesiek, Urban, Hubert, Schwarz, Maier-Hein, Bendszus, and Biller}{Kleesiek et~al\mbox{.}}{2016}]%
        {kleesiek_deep_2016}
\bibfield{author}{\bibinfo{person}{Jens Kleesiek}, \bibinfo{person}{Gregor Urban}, \bibinfo{person}{Alexander Hubert}, \bibinfo{person}{Daniel Schwarz}, \bibinfo{person}{Klaus Maier-Hein}, \bibinfo{person}{Martin Bendszus}, {and} \bibinfo{person}{Armin Biller}.} \bibinfo{year}{2016}\natexlab{}.
\newblock \showarticletitle{Deep {MRI} brain extraction: {A} {3D} convolutional neural network for skull stripping}.
\newblock \bibinfo{journal}{\emph{NeuroImage}}  \bibinfo{volume}{129} (\bibinfo{year}{2016}), \bibinfo{pages}{460 -- 469}.
\newblock
\urldef\tempurl%
\url{https://doi.org/10.1016/j.neuroimage.2016.01.024}
\showDOI{\tempurl}


\bibitem[\protect\citeauthoryear{Kolter and Maloof}{Kolter and Maloof}{2007}]%
        {kolter_dynamic_2007}
\bibfield{author}{\bibinfo{person}{J.~Zico Kolter} {and} \bibinfo{person}{Marcus~A. Maloof}.} \bibinfo{year}{2007}\natexlab{}.
\newblock \showarticletitle{Dynamic weighted majority: {An} ensemble method for drifting concepts}.
\newblock \bibinfo{journal}{\emph{Journal of Machine Learning Research}}  \bibinfo{volume}{8} (\bibinfo{year}{2007}), \bibinfo{pages}{2755 -- 2790}.
\newblock


\bibitem[\protect\citeauthoryear{Krawczyk, Minku, Gama, Stefanowski, and Wo{\a'z}niak}{Krawczyk et~al\mbox{.}}{2017}]%
        {krawczyk_ensemble_2017}
\bibfield{author}{\bibinfo{person}{Bartosz Krawczyk}, \bibinfo{person}{Leandro~L. Minku}, \bibinfo{person}{Jo{\~a}o Gama}, \bibinfo{person}{Jerzy Stefanowski}, {and} \bibinfo{person}{Micha{\l} Wo{\a'z}niak}.} \bibinfo{year}{2017}\natexlab{}.
\newblock \showarticletitle{Ensemble learning for data stream analysis: {A} survey}.
\newblock \bibinfo{journal}{\emph{Information Fusion}}  \bibinfo{volume}{37} (\bibinfo{year}{2017}), \bibinfo{pages}{132 -- 156}.
\newblock
\urldef\tempurl%
\url{https://doi.org/10.1016/j.inffus.2017.02.004}
\showDOI{\tempurl}


\bibitem[\protect\citeauthoryear{Kumar, McCann, Naughton, and Patel}{Kumar et~al\mbox{.}}{2016}]%
        {kumar_model_2016}
\bibfield{author}{\bibinfo{person}{Arun Kumar}, \bibinfo{person}{Robert McCann}, \bibinfo{person}{Jeffrey Naughton}, {and} \bibinfo{person}{Jignesh~M. Patel}.} \bibinfo{year}{2016}\natexlab{}.
\newblock \showarticletitle{Model {Selection} {Management} {Systems}: {The} {Next} {Frontier} of {Advanced} {Analytics}}.
\newblock \bibinfo{journal}{\emph{ACM SIGMOD Record}} \bibinfo{volume}{44}, \bibinfo{number}{4} (\bibinfo{date}{may} \bibinfo{year}{2016}), \bibinfo{pages}{17--22}.
\newblock
\showISSN{0163-5808}
\urldef\tempurl%
\url{https://doi.org/10.1145/2935694.2935698}
\showDOI{\tempurl}


\bibitem[\protect\citeauthoryear{Lones}{Lones}{2011}]%
        {lones_sean_2011}
\bibfield{author}{\bibinfo{person}{Michael Lones}.} \bibinfo{year}{2011}\natexlab{}.
\newblock \showarticletitle{Sean {Luke}: essentials of metaheuristics}.
\newblock \bibinfo{journal}{\emph{Genetic Programming and Evolvable Machines}} \bibinfo{volume}{12}, \bibinfo{number}{3} (\bibinfo{date}{sep} \bibinfo{year}{2011}), \bibinfo{pages}{333--334}.
\newblock
\showISSN{1573-7632}
\urldef\tempurl%
\url{https://doi.org/10.1007/s10710-011-9139-0}
\showDOI{\tempurl}


\bibitem[\protect\citeauthoryear{Lorbeer, Kosareva, Deva, Softi{\a'c}, Ruppel, and K{\"u}pper}{Lorbeer et~al\mbox{.}}{2018}]%
        {lorbeer_variations_2018}
\bibfield{author}{\bibinfo{person}{Boris Lorbeer}, \bibinfo{person}{Ana Kosareva}, \bibinfo{person}{Bersant Deva}, \bibinfo{person}{D{\v z}enan Softi{\a'c}}, \bibinfo{person}{Peter Ruppel}, {and} \bibinfo{person}{Axel K{\"u}pper}.} \bibinfo{year}{2018}\natexlab{}.
\newblock \showarticletitle{Variations on the {Clustering} {Algorithm} {BIRCH}}.
\newblock \bibinfo{journal}{\emph{Big Data Research}}  \bibinfo{volume}{11} (\bibinfo{year}{2018}), \bibinfo{pages}{44 -- 53}.
\newblock
\urldef\tempurl%
\url{https://doi.org/10.1016/j.bdr.2017.09.002}
\showDOI{\tempurl}


\bibitem[\protect\citeauthoryear{Lu, Liu, Dong, Gu, Gama, and Zhang}{Lu et~al\mbox{.}}{2019}]%
        {lu_learning_2019}
\bibfield{author}{\bibinfo{person}{Jie Lu}, \bibinfo{person}{Anjin Liu}, \bibinfo{person}{Fan Dong}, \bibinfo{person}{Feng Gu}, \bibinfo{person}{Joao Gama}, {and} \bibinfo{person}{Guangquan Zhang}.} \bibinfo{year}{2019}\natexlab{}.
\newblock \showarticletitle{Learning under {Concept} {Drift}: {A} {Review}}.
\newblock \bibinfo{journal}{\emph{IEEE Transactions on Knowledge and Data Engineering}} \bibinfo{volume}{31}, \bibinfo{number}{12} (\bibinfo{year}{2019}), \bibinfo{pages}{2346 -- 2363}.
\newblock
\urldef\tempurl%
\url{https://doi.org/10.1109/TKDE.2018.2876857}
\showDOI{\tempurl}


\bibitem[\protect\citeauthoryear{Luong, Nguyen, Liew, and Wang}{Luong et~al\mbox{.}}{2021}]%
        {luong_heterogeneous_2021}
\bibfield{author}{\bibinfo{person}{Anh~Vu Luong}, \bibinfo{person}{Tien~Thanh Nguyen}, \bibinfo{person}{Alan Wee-Chung Liew}, {and} \bibinfo{person}{Shilin Wang}.} \bibinfo{year}{2021}\natexlab{}.
\newblock \showarticletitle{Heterogeneous ensemble selection for evolving data streams}.
\newblock \bibinfo{journal}{\emph{Pattern Recognition}}  \bibinfo{volume}{112} (\bibinfo{year}{2021}).
\newblock
\urldef\tempurl%
\url{https://doi.org/10.1016/j.patcog.2020.107743}
\showDOI{\tempurl}


\bibitem[\protect\citeauthoryear{Matsubara, Sakurai, {Van Panhuis}, and Faloutsos}{Matsubara et~al\mbox{.}}{2014}]%
        {matsubara_funnel_2014}
\bibfield{author}{\bibinfo{person}{Yasuko Matsubara}, \bibinfo{person}{Yasushi Sakurai}, \bibinfo{person}{Willem~G. {Van Panhuis}}, {and} \bibinfo{person}{Christos Faloutsos}.} \bibinfo{year}{2014}\natexlab{}.
\newblock \showarticletitle{{FUNNEL}: {Automatic} mining of spatially coevolving epidemics}. In \bibinfo{booktitle}{\emph{Proceedings of the {ACM} {SIGKDD} {International} {Conference} on {Knowledge} {Discovery} and {Data} {Mining}}}. \bibinfo{publisher}{Association for Computing Machinery}, \bibinfo{pages}{105 -- 114}.
\newblock
\showISBNx{978-1-4503-2956-9}
\urldef\tempurl%
\url{https://doi.org/10.1145/2623330.2623624}
\showDOI{\tempurl}


\bibitem[\protect\citeauthoryear{Mendes-Moreira, Soares, Jorge, and {De Sousa}}{Mendes-Moreira et~al\mbox{.}}{2012}]%
        {mendes-moreira_ensemble_2012}
\bibfield{author}{\bibinfo{person}{Jo{\~a}o Mendes-Moreira}, \bibinfo{person}{Carlos Soares}, \bibinfo{person}{A{\a'l}ipio~M{\a'a}rio Jorge}, {and} \bibinfo{person}{Jorge~Freire {De Sousa}}.} \bibinfo{year}{2012}\natexlab{}.
\newblock \showarticletitle{Ensemble approaches for regression: {A} survey}.
\newblock \bibinfo{journal}{\emph{Comput. Surveys}} \bibinfo{volume}{45}, \bibinfo{number}{1} (\bibinfo{year}{2012}).
\newblock
\urldef\tempurl%
\url{https://doi.org/10.1145/2379776.2379786}
\showDOI{\tempurl}


\bibitem[\protect\citeauthoryear{M{\"o}ller-Levet, Klawonn, Cho, and Wolkenhauer}{M{\"o}ller-Levet et~al\mbox{.}}{2003}]%
        {moller-levet_fuzzy_2003}
\bibfield{author}{\bibinfo{person}{Carla~S. M{\"o}ller-Levet}, \bibinfo{person}{Frank Klawonn}, \bibinfo{person}{Kwang-Hyun Cho}, {and} \bibinfo{person}{Olaf Wolkenhauer}.} \bibinfo{year}{2003}\natexlab{}.
\newblock \showarticletitle{Fuzzy clustering of short time-series and unevenly distributed sampling points}. In \bibinfo{booktitle}{\emph{International symposium on intelligent data analysis}}, \bibfield{editor}{\bibinfo{person}{Berthold M.R}, \bibinfo{person}{Lenz H.-J}, \bibinfo{person}{Bradley E}, \bibinfo{person}{Kruse R}, {and} \bibinfo{person}{Borgelt C}} (Eds.), Vol.~\bibinfo{volume}{2810}. \bibinfo{publisher}{Springer}, \bibinfo{pages}{330 -- 340}.
\newblock
\urldef\tempurl%
\url{https://doi.org/10.1007/978-3-540-45231-7_31}
\showDOI{\tempurl}


\bibitem[\protect\citeauthoryear{Nakandala, Zhang, and Kumar}{Nakandala et~al\mbox{.}}{2020}]%
        {nakandala_cerebro_2020}
\bibfield{author}{\bibinfo{person}{Supun Nakandala}, \bibinfo{person}{Yuhao Zhang}, {and} \bibinfo{person}{Arun Kumar}.} \bibinfo{year}{2020}\natexlab{}.
\newblock \showarticletitle{Cerebro: {A} data system for optimized deep learning model selection}.
\newblock \bibinfo{journal}{\emph{Proceedings of the VLDB Endowment}} \bibinfo{volume}{13}, \bibinfo{number}{11} (\bibinfo{year}{2020}), \bibinfo{pages}{2159 -- 2173}.
\newblock
\urldef\tempurl%
\url{https://doi.org/10.14778/3407790.3407816}
\showDOI{\tempurl}


\bibitem[\protect\citeauthoryear{Nie, Zhang, Adeli, Liu, and Shen}{Nie et~al\mbox{.}}{2016}]%
        {nie_3d_2016}
\bibfield{author}{\bibinfo{person}{Dong Nie}, \bibinfo{person}{Han Zhang}, \bibinfo{person}{Ehsan Adeli}, \bibinfo{person}{Luyan Liu}, {and} \bibinfo{person}{Dinggang Shen}.} \bibinfo{year}{2016}\natexlab{}.
\newblock \showarticletitle{{3D} deep learning for multi-modal imaging-guided survival time prediction of brain tumor patients}. In \bibinfo{booktitle}{\emph{Lecture {Notes} in {Computer} {Science} (including subseries {Lecture} {Notes} in {Artificial} {Intelligence} and {Lecture} {Notes} in {Bioinformatics})}}, \bibfield{editor}{\bibinfo{person}{Unal G}, \bibinfo{person}{Ourselin S}, \bibinfo{person}{Joskowicz L}, \bibinfo{person}{Sabuncu M.R}, {and} \bibinfo{person}{Wells W}} (Eds.), Vol.~\bibinfo{volume}{9901 LNCS}. \bibinfo{publisher}{Springer}, \bibinfo{pages}{212 -- 220}.
\newblock
\urldef\tempurl%
\url{https://doi.org/10.1007/978-3-319-46723-8_25}
\showDOI{\tempurl}


\bibitem[\protect\citeauthoryear{Pavlyshenko}{Pavlyshenko}{2018}]%
        {pavlyshenko_using_2018}
\bibfield{author}{\bibinfo{person}{Bohdan Pavlyshenko}.} \bibinfo{year}{2018}\natexlab{}.
\newblock \showarticletitle{Using {Stacking} {Approaches} for {Machine} {Learning} {Models}}. In \bibinfo{booktitle}{\emph{Proceedings of the 2018 {IEEE} 2nd {International} {Conference} on {Data} {Stream} {Mining} and {Processing}, {DSMP} 2018}}. \bibinfo{publisher}{IEEE}, \bibinfo{pages}{255 -- 258}.
\newblock
\showISBNx{978-1-5386-2874-4}
\urldef\tempurl%
\url{https://doi.org/10.1109/DSMP.2018.8478522}
\showDOI{\tempurl}


\bibitem[\protect\citeauthoryear{Pereira, Souto, Chaves, Zorilla, Tsan, Rusu, Ogasawara, Ziviani, and Porto}{Pereira et~al\mbox{.}}{2021}]%
        {pereira_djensemble_2021}
\bibfield{author}{\bibinfo{person}{Rafael Pereira}, \bibinfo{person}{Yania Souto}, \bibinfo{person}{Anderson Chaves}, \bibinfo{person}{Rocio Zorilla}, \bibinfo{person}{Brian Tsan}, \bibinfo{person}{Florin Rusu}, \bibinfo{person}{Eduardo Ogasawara}, \bibinfo{person}{Artur Ziviani}, {and} \bibinfo{person}{Fabio Porto}.} \bibinfo{year}{2021}\natexlab{}.
\newblock \showarticletitle{{DJEnsemble}: {A} {Cost}-{Based} {Selection} and {Allocation} of a {Disjoint} {Ensemble} of {Spatio}-{Temporal} {Models}}. In \bibinfo{booktitle}{\emph{{ACM} {International} {Conference} {Proceeding} {Series}}}, \bibfield{editor}{\bibinfo{person}{Zhu Q}, \bibinfo{person}{Zhu X}, \bibinfo{person}{Tu~Y}, \bibinfo{person}{Xu~Z}, {and} \bibinfo{person}{Kumar A}} (Eds.). \bibinfo{publisher}{Association for Computing Machinery}, \bibinfo{pages}{226 -- 231}.
\newblock
\showISBNx{978-1-4503-8413-1}
\urldef\tempurl%
\url{https://doi.org/10.1145/3468791.3468806}
\showDOI{\tempurl}


\bibitem[\protect\citeauthoryear{Polikar, Udpa, Udpa, and Honavar}{Polikar et~al\mbox{.}}{2001}]%
        {polikar_learn_2001}
\bibfield{author}{\bibinfo{person}{Robi Polikar}, \bibinfo{person}{Lalita Udpa}, \bibinfo{person}{Satish~S. Udpa}, {and} \bibinfo{person}{Vasant Honavar}.} \bibinfo{year}{2001}\natexlab{}.
\newblock \showarticletitle{Learn++: {An} incremental learning algorithm for supervised neural networks}.
\newblock \bibinfo{journal}{\emph{IEEE Transactions on Systems, Man and Cybernetics Part C: Applications and Reviews}} \bibinfo{volume}{31}, \bibinfo{number}{4} (\bibinfo{year}{2001}), \bibinfo{pages}{497 -- 508}.
\newblock
\urldef\tempurl%
\url{https://doi.org/10.1109/5326.983933}
\showDOI{\tempurl}


\bibitem[\protect\citeauthoryear{Ramberg and Schmeiser}{Ramberg and Schmeiser}{1974}]%
        {ramberg_approximate_1974}
\bibfield{author}{\bibinfo{person}{John~S. Ramberg} {and} \bibinfo{person}{Bruce~W. Schmeiser}.} \bibinfo{year}{1974}\natexlab{}.
\newblock \showarticletitle{An approximate method for generating asymmetric random variables}.
\newblock \bibinfo{journal}{\emph{Commun. ACM}} \bibinfo{volume}{17}, \bibinfo{number}{2} (\bibinfo{year}{1974}), \bibinfo{pages}{78 -- 82}.
\newblock
\urldef\tempurl%
\url{https://doi.org/10.1145/360827.360840}
\showDOI{\tempurl}


\bibitem[\protect\citeauthoryear{Ravuri, Lenc, Willson, Kangin, Lam, Mirowski, Fitzsimons, Athanassiadou, Kashem, Madge, Prudden, Mandhane, Clark, Brock, Simonyan, Hadsell, Robinson, Clancy, Arribas, and Mohamed}{Ravuri et~al\mbox{.}}{2021}]%
        {ravuri_skilful_2021}
\bibfield{author}{\bibinfo{person}{Suman Ravuri}, \bibinfo{person}{Karel Lenc}, \bibinfo{person}{Matthew Willson}, \bibinfo{person}{Dmitry Kangin}, \bibinfo{person}{Remi Lam}, \bibinfo{person}{Piotr Mirowski}, \bibinfo{person}{Megan Fitzsimons}, \bibinfo{person}{Maria Athanassiadou}, \bibinfo{person}{Sheleem Kashem}, \bibinfo{person}{Sam Madge}, \bibinfo{person}{Rachel Prudden}, \bibinfo{person}{Amol Mandhane}, \bibinfo{person}{Aidan Clark}, \bibinfo{person}{Andrew Brock}, \bibinfo{person}{Karen Simonyan}, \bibinfo{person}{Raia Hadsell}, \bibinfo{person}{Niall Robinson}, \bibinfo{person}{Ellen Clancy}, \bibinfo{person}{Alberto Arribas}, {and} \bibinfo{person}{Shakir Mohamed}.} \bibinfo{year}{2021}\natexlab{}.
\newblock \showarticletitle{Skilful precipitation nowcasting using deep generative models of radar}.
\newblock \bibinfo{journal}{\emph{Nature}} \bibinfo{volume}{597}, \bibinfo{number}{7878} (\bibinfo{year}{2021}), \bibinfo{pages}{672 -- 677}.
\newblock
\urldef\tempurl%
\url{https://doi.org/10.1038/s41586-021-03854-z}
\showDOI{\tempurl}


\bibitem[\protect\citeauthoryear{Saffari, Leistner, Santner, Godec, and Bischof}{Saffari et~al\mbox{.}}{2009}]%
        {saffari_-line_2009}
\bibfield{author}{\bibinfo{person}{Amir Saffari}, \bibinfo{person}{Christian Leistner}, \bibinfo{person}{Jakob Santner}, \bibinfo{person}{Martin Godec}, {and} \bibinfo{person}{Horst Bischof}.} \bibinfo{year}{2009}\natexlab{}.
\newblock \showarticletitle{On-line random forests}. In \bibinfo{booktitle}{\emph{2009 {IEEE} 12th {International} {Conference} on {Computer} {Vision} {Workshops}, {ICCV} {Workshops} 2009}}. \bibinfo{publisher}{IEEE}, \bibinfo{pages}{1393 -- 1400}.
\newblock
\showISBNx{978-1-4244-4442-7}
\urldef\tempurl%
\url{https://doi.org/10.1109/ICCVW.2009.5457447}
\showDOI{\tempurl}


\bibitem[\protect\citeauthoryear{Sagi and Rokach}{Sagi and Rokach}{2018}]%
        {sagi_ensemble_2018}
\bibfield{author}{\bibinfo{person}{Omer Sagi} {and} \bibinfo{person}{Lior Rokach}.} \bibinfo{year}{2018}\natexlab{}.
\newblock \showarticletitle{Ensemble learning: {A} survey}.
\newblock \bibinfo{journal}{\emph{Wiley Interdisciplinary Reviews: Data Mining and Knowledge Discovery}} \bibinfo{volume}{8}, \bibinfo{number}{4} (\bibinfo{year}{2018}).
\newblock
\urldef\tempurl%
\url{https://doi.org/10.1002/widm.1249}
\showDOI{\tempurl}


\bibitem[\protect\citeauthoryear{Sagl, Resch, Hawelka, and Beinat}{Sagl et~al\mbox{.}}{2012}]%
        {sagl_social_2012}
\bibfield{author}{\bibinfo{person}{G{\"u}nther Sagl}, \bibinfo{person}{Bernd Resch}, \bibinfo{person}{Bartosz Hawelka}, {and} \bibinfo{person}{Euro Beinat}.} \bibinfo{year}{2012}\natexlab{}.
\newblock \showarticletitle{From {Social} {Sensor} {Data} to {Collective} {Human} {Behaviour} {Patterns}: {Analysing} and {Visualising} {Spatio}-{Temporal} {Dynamics} in {Urban} {Environments}}. In \bibinfo{booktitle}{\emph{Proceedings of the {GI}-{Forum}}}. \bibinfo{publisher}{Herbert Wichmann Verlag Berlin}, \bibinfo{pages}{54--63}.
\newblock


\bibitem[\protect\citeauthoryear{Schelter, Biessmann, Januschowski, Salinas, Seufert, and Szarvas}{Schelter et~al\mbox{.}}{2018}]%
        {schelter_challenges_2018}
\bibfield{author}{\bibinfo{person}{Sebastian Schelter}, \bibinfo{person}{Felix Biessmann}, \bibinfo{person}{Tim Januschowski}, \bibinfo{person}{David Salinas}, \bibinfo{person}{Stephan Seufert}, {and} \bibinfo{person}{Gyuri Szarvas}.} \bibinfo{year}{2018}\natexlab{}.
\newblock \bibinfo{booktitle}{\emph{On challenges in machine learning model management}}.
\newblock \bibinfo{type}{{T}echnical {R}eport}. \bibinfo{institution}{https://www.amazon.science/publications/on-challenges-in-machine-learning-model-management}. \bibinfo{pages}{5--15} pages.
\newblock


\bibitem[\protect\citeauthoryear{Shi, Chen, Wang, Yeung, Wong, and Woo}{Shi et~al\mbox{.}}{2015}]%
        {shi_convolutional_2015}
\bibfield{author}{\bibinfo{person}{Xingjian Shi}, \bibinfo{person}{Zhourong Chen}, \bibinfo{person}{Hao Wang}, \bibinfo{person}{Dit-Yan Yeung}, \bibinfo{person}{Wai-Kin Wong}, {and} \bibinfo{person}{Wang-Chun Woo}.} \bibinfo{year}{2015}\natexlab{}.
\newblock \showarticletitle{Convolutional {LSTM} network: {A} machine learning approach for precipitation nowcasting}. In \bibinfo{booktitle}{\emph{Advances in {Neural} {Information} {Processing} {Systems}}} \emph{(\bibinfo{series}{{NIPS}'15})}, Vol.~\bibinfo{volume}{2015-January}. \bibinfo{publisher}{MIT Press}, \bibinfo{address}{Cambridge, MA, USA}, \bibinfo{pages}{802 -- 810}.
\newblock


\bibitem[\protect\citeauthoryear{Shi and Yeung}{Shi and Yeung}{2018}]%
        {shi_machine_2018}
\bibfield{author}{\bibinfo{person}{Xingjian Shi} {and} \bibinfo{person}{Dit-Yan Yeung}.} \bibinfo{year}{2018}\natexlab{}.
\newblock \bibinfo{title}{Machine {Learning} for {Spatiotemporal} {Sequence} {Forecasting}: {A} {Survey}}.
\newblock
\newblock
\urldef\tempurl%
\url{https://doi.org/10.48550/arXiv.1808.06865}
\showDOI{\tempurl}


\bibitem[\protect\citeauthoryear{Sun, Dubey, and White}{Sun et~al\mbox{.}}{2017}]%
        {sun_dxnat_2017}
\bibfield{author}{\bibinfo{person}{Fangzhou Sun}, \bibinfo{person}{Abhishek Dubey}, {and} \bibinfo{person}{Jules White}.} \bibinfo{year}{2017}\natexlab{}.
\newblock \showarticletitle{{DxNAT} - {Deep} neural networks for explaining non-recurring traffic congestion}. In \bibinfo{booktitle}{\emph{Proceedings - 2017 {IEEE} {International} {Conference} on {Big} {Data}, {Big} {Data} 2017}}, \bibfield{editor}{\bibinfo{person}{Nie J.-Y}, \bibinfo{person}{Obradovic Z}, \bibinfo{person}{Suzumura T}, \bibinfo{person}{Ghosh R}, \bibinfo{person}{Nambiar R}, \bibinfo{person}{Wang C}, \bibinfo{person}{Zang H}, \bibinfo{person}{Baeza-Yates R}, \bibinfo{person}{Baeza-Yates R}, \bibinfo{person}{Hu~X}, \bibinfo{person}{Kepner J}, \bibinfo{person}{Cuzzocrea A}, \bibinfo{person}{Tang J}, {and} \bibinfo{person}{Toyoda M}} (Eds.), Vol.~\bibinfo{volume}{2018-January}. \bibinfo{publisher}{IEEE}, \bibinfo{pages}{2141 -- 2150}.
\newblock
\showISBNx{978-1-5386-2714-3}
\urldef\tempurl%
\url{https://doi.org/10.1109/BigData.2017.8258162}
\showDOI{\tempurl}


\bibitem[\protect\citeauthoryear{{van Staden} and Loots}{{van Staden} and Loots}{2009}]%
        {staden_method_2009}
\bibfield{author}{\bibinfo{person}{Paul~J. {van Staden}} {and} \bibinfo{person}{M.~T. Loots}.} \bibinfo{year}{2009}\natexlab{}.
\newblock \showarticletitle{Method of {L}-moment estimation for the generalized lambda distribution}. In \bibinfo{booktitle}{\emph{Proceedings of the {Third} {Annual} {ASEARC} {Conference}}}. \bibinfo{pages}{7--8}.
\newblock


\bibitem[\protect\citeauthoryear{Wang, Gao, Zhang, Wang, Chen, Ng, Ooi, Shao, and Reyad}{Wang et~al\mbox{.}}{2018}]%
        {wang_rafiki_2018}
\bibfield{author}{\bibinfo{person}{Wei Wang}, \bibinfo{person}{Jinyang Gao}, \bibinfo{person}{Meihui Zhang}, \bibinfo{person}{Sheng Wang}, \bibinfo{person}{Gang Chen}, \bibinfo{person}{Teck~Khim Ng}, \bibinfo{person}{Beng~Chin Ooi}, \bibinfo{person}{Jie Shao}, {and} \bibinfo{person}{Moaz Reyad}.} \bibinfo{year}{2018}\natexlab{}.
\newblock \showarticletitle{Rafiki: {Machine} learning as an analytics service system}. In \bibinfo{booktitle}{\emph{Proceedings of the {VLDB} {Endowment}}}, Vol.~\bibinfo{volume}{12}. \bibinfo{publisher}{Association for Computing Machinery}, \bibinfo{pages}{128 -- 140}.
\newblock
\urldef\tempurl%
\url{https://doi.org/10.14778/3282495.3282499}
\showDOI{\tempurl}


\bibitem[\protect\citeauthoryear{{Warren Liao}}{{Warren Liao}}{2005}]%
        {warren_liao_clustering_2005}
\bibfield{author}{\bibinfo{person}{T. {Warren Liao}}.} \bibinfo{year}{2005}\natexlab{}.
\newblock \showarticletitle{Clustering of time series data - {A} survey}.
\newblock \bibinfo{journal}{\emph{Pattern Recognition}} \bibinfo{volume}{38}, \bibinfo{number}{11} (\bibinfo{year}{2005}), \bibinfo{pages}{1857 -- 1874}.
\newblock
\urldef\tempurl%
\url{https://doi.org/10.1016/j.patcog.2005.01.025}
\showDOI{\tempurl}


\bibitem[\protect\citeauthoryear{Wen, Wei, Zhou, Li, Zhang, and Han}{Wen et~al\mbox{.}}{2018}]%
        {wen_deep_2018}
\bibfield{author}{\bibinfo{person}{Dong Wen}, \bibinfo{person}{Zhenhao Wei}, \bibinfo{person}{Yanhong Zhou}, \bibinfo{person}{Guolin Li}, \bibinfo{person}{Xu Zhang}, {and} \bibinfo{person}{Wei Han}.} \bibinfo{year}{2018}\natexlab{}.
\newblock \showarticletitle{Deep learning methods to process fmri data and their application in the diagnosis of cognitive impairment: {A} brief overview and our opinion}.
\newblock \bibinfo{journal}{\emph{Frontiers in Neuroinformatics}}  \bibinfo{volume}{12} (\bibinfo{year}{2018}).
\newblock
\urldef\tempurl%
\url{https://doi.org/10.3389/fninf.2018.00023}
\showDOI{\tempurl}


\bibitem[\protect\citeauthoryear{Wolpert}{Wolpert}{1992}]%
        {wolpert_stacked_1992}
\bibfield{author}{\bibinfo{person}{David~H. Wolpert}.} \bibinfo{year}{1992}\natexlab{}.
\newblock \showarticletitle{Stacked generalization}.
\newblock \bibinfo{journal}{\emph{Neural Networks}} \bibinfo{volume}{5}, \bibinfo{number}{2} (\bibinfo{year}{1992}), \bibinfo{pages}{241 -- 259}.
\newblock
\urldef\tempurl%
\url{https://doi.org/10.1016/S0893-6080(05)80023-1}
\showDOI{\tempurl}


\bibitem[\protect\citeauthoryear{Wolpert}{Wolpert}{2002}]%
        {wolpert_supervised_2002}
\bibfield{author}{\bibinfo{person}{David~H. Wolpert}.} \bibinfo{year}{2002}\natexlab{}.
\newblock \showarticletitle{The {Supervised} {Learning} {No}-{Free}-{Lunch} {Theorems}}.
\newblock \bibinfo{journal}{\emph{Soft Computing and Industry: Recent Applications}} (\bibinfo{year}{2002}), \bibinfo{pages}{25--42}.
\newblock
\urldef\tempurl%
\url{https://doi.org/10.1007/978-1-4471-0123-9_3}
\showDOI{\tempurl}


\bibitem[\protect\citeauthoryear{Yagoubi, Akbarinia, Kolev, Levchenko, Masseglia, Valduriez, and Shasha}{Yagoubi et~al\mbox{.}}{2018}]%
        {yagoubi_parcorr_2018}
\bibfield{author}{\bibinfo{person}{Djamel~Edine Yagoubi}, \bibinfo{person}{Reza Akbarinia}, \bibinfo{person}{Boyan Kolev}, \bibinfo{person}{Oleksandra Levchenko}, \bibinfo{person}{Florent Masseglia}, \bibinfo{person}{Patrick Valduriez}, {and} \bibinfo{person}{Dennis Shasha}.} \bibinfo{year}{2018}\natexlab{}.
\newblock \showarticletitle{{ParCorr}: efficient parallel methods to identify similar time series pairs across sliding windows}.
\newblock \bibinfo{journal}{\emph{Data Mining and Knowledge Discovery}} \bibinfo{volume}{32}, \bibinfo{number}{5} (\bibinfo{year}{2018}), \bibinfo{pages}{1481 -- 1507}.
\newblock
\urldef\tempurl%
\url{https://doi.org/10.1007/s10618-018-0580-z}
\showDOI{\tempurl}


\bibitem[\protect\citeauthoryear{Yu, Xia, Li, Hou, and Sheng}{Yu et~al\mbox{.}}{2023}]%
        {yu_spatio-temporal_2023}
\bibfield{author}{\bibinfo{person}{Shuo Yu}, \bibinfo{person}{Feng Xia}, \bibinfo{person}{Shihao Li}, \bibinfo{person}{Mingliang Hou}, {and} \bibinfo{person}{Quan~Z. Sheng}.} \bibinfo{year}{2023}\natexlab{}.
\newblock \showarticletitle{Spatio-temporal {Graph} {Learning} for {Epidemic} {Prediction}}.
\newblock \bibinfo{journal}{\emph{ACM Transactions on Intelligent Systems and Technology}} \bibinfo{volume}{14}, \bibinfo{number}{2} (\bibinfo{year}{2023}).
\newblock
\urldef\tempurl%
\url{https://doi.org/10.1145/3579815}
\showDOI{\tempurl}


\bibitem[\protect\citeauthoryear{Zhang, Ramakrishnan, and Livny}{Zhang et~al\mbox{.}}{1996}]%
        {zhang_birch_1996}
\bibfield{author}{\bibinfo{person}{Tian Zhang}, \bibinfo{person}{Raghu Ramakrishnan}, {and} \bibinfo{person}{Miron Livny}.} \bibinfo{year}{1996}\natexlab{}.
\newblock \showarticletitle{{BIRCH}: {An} {Efficient} {Data} {Clustering} {Method} for {Very} {Large} {Databases}}.
\newblock \bibinfo{journal}{\emph{SIGMOD Record (ACM Special Interest Group on Management of Data)}} \bibinfo{volume}{25}, \bibinfo{number}{2} (\bibinfo{year}{1996}), \bibinfo{pages}{103 -- 114}.
\newblock
\urldef\tempurl%
\url{https://doi.org/10.1145/235968.233324}
\showDOI{\tempurl}


\end{thebibliography}

\end{document}